\documentclass[pdflatex, sn-basic, iicol]{sn-jnl}

\usepackage{graphicx}
\usepackage{subcaption}
\usepackage{amsmath}
\usepackage{framed}
\usepackage{appendix}
\usepackage{multicol}
\usepackage{stfloats}
\usepackage{balance}
\usepackage[hang,flushmargin]{footmisc} 

\newsavebox\CBox
\def\textBF#1{\sbox\CBox{#1}\resizebox{\wd\CBox}{\ht\CBox}{\textbf{#1}}}
\jyear{2021}%

\theoremstyle{thmstyleone}%
\theoremstyle{thmstyletwo}%

\theoremstyle{thmstylethree}%

\begin{document}

\title[Ege Berkay Gulcan, Fazli Can]{Implicit Concept Drift Detection for Multi-label Data Streams}

\author{\fnm{Ege Berkay} \sur{Gulcan}}\email{berkay.gulcan@bilkent.edu.tr}

\author{\fnm{Fazli} \sur{Can}}\email{canf@cs.bilkent.edu.tr}

\affil{\orgdiv{Bilkent Information Retrieval Group}, \orgname{Computer Engineering Department, Bilkent University}, \orgaddress{\city{Ankara}, \country{Turkey}}}

\abstract{Many real-world applications adopt multi-label data streams as the need for algorithms to deal with rapidly changing data increases. Changes in data distribution, also known as concept drift, cause the existing classification models to rapidly lose their effectiveness. To assist the classifiers, we propose a novel algorithm called Label Dependency Drift Detector (LD3), an implicit (unsupervised) concept drift detector using label dependencies within the data for multi-label data streams. Our study exploits the dynamic temporal dependencies between labels using a label influence ranking method, which leverages a data fusion algorithm and uses the produced ranking to detect concept drift. LD3 is the first unsupervised concept drift detection algorithm in the multi-label classification problem area. In this study, we perform an extensive evaluation of LD3 by comparing it with 14 prevalent supervised concept drift detection algorithms that we adapt to the problem area using 12 datasets and a baseline classifier. The results show that LD3 provides between 19.8\% and 68.6\% better predictive performance than comparable detectors on both real-world and synthetic data streams.}

\renewcommand{\thefootnote}{\arabic{footnote}}


\maketitle
\section{Introduction}\label{sec1}

Many organizations generate temporal data in the form of data streams with high variety, volume, and velocity \citep{zheng2019survey}. Data is created continuously on a massive scale and can be assigned multiple labels, which is termed multi-labeled. 
However, because of the scale of this data, they need to be processed immediately since the cost associated with storage and retrieval is high, which explains the current popularity of data stream mining \citep{bahri2021data}.

\par
Real-world applications are constantly evolving and over time, a change in data distribution may occur. This change in data is called concept drift \citep{bonab2018goowe} which is one of the most prevalent problems in data stream mining. Many of the traditional classification algorithms assume that the data is static, which causes them to lose effectiveness when faced with concept drift. An example area for concept drift is the energy sector where the drifted streams may cause instabilities for the learning models designed to predict energy consumption, production and distribution \citep{hammami2020neural}. 

\par
In this study, we propose a novel unsupervised (implicit) concept drift detection algorithm that exploits label dependencies between class labels for multi-label data streams through data fusion methods. In multi-label data stream mining, it is common for labels to have correlations and dependencies \citep{xu2019survey}. Several studies \citep{guo2011multi, wang2016cnn, zhang2010multi} show that incorporating label dependencies into multi-label classifiers boosts their effectiveness. We aim to utilize these correlations among labels to demonstrate that label dependencies can be used for detecting concept drift. For this purpose, we model the label dependencies by creating a co-occurrence matrix of labels \citep{xue2011correlative}, and we use the generated matrix to represent the current and past stream representation through label ranking and then use data fusion to detect concept drift.
\par
In this study our main contributions are the following.
\begin{enumerate}
    \item To the best of our knowledge, for the first time in literature, we introduce the concept of label dependency ranking and use it for concept drift detection in multi-label classification.
    \item We perform an extensive evaluation of our method by comparing it with 14 prevalent concept drift detection algorithms using 12 data streams and a base classifier. We compare drift detection algorithms with their influence on the predictive performance of the baseline classifier. In all cases, LD3 is the number one algorithm and in most cases, it enables performance that is statistically significantly higher than most of the baselines in terms of almost all effectiveness measures utilized in the experiments. Furthermore, our work provides the first study on the use of several different concept drift detection algorithms which are developed for multi-class environments, for concept drift detection in the multi-label classification problem domain.
    \item Our method LD3 is the first unsupervised concept drift detection algorithm in the problem domain we focus on. The baseline drift detection algorithms used in the experiments are supervised and require true labels; on the other hand, our method uses only the predicted labels: In many streaming environments, true class labels needed by supervised methods are not always available; in some cases, only a percentage of them are available or arrive late or are potentially unavailable \citep{sethi2017reliable, vzliobaite2010change}. These factors show the practical importance of our approach.

\end{enumerate}

\begin{figure*}[bp]
\centering
\begin{subfigure}{.7\columnwidth}
  \centering
  \includegraphics[width=.7\columnwidth]{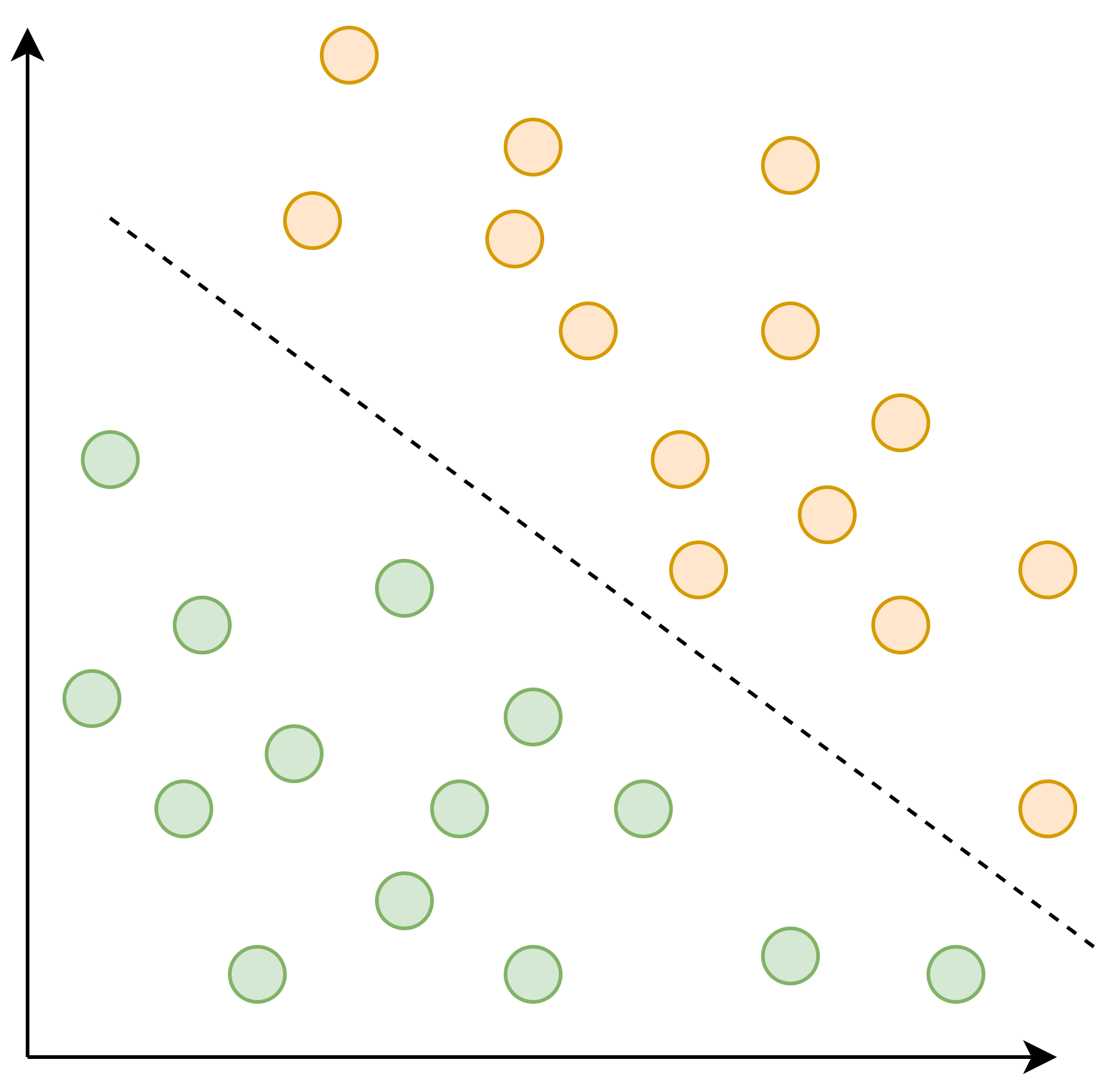}
  \caption{Original data}
  \label{fig:sub1}
\end{subfigure}%
\begin{subfigure}{.7\columnwidth}
  \centering
  \includegraphics[width=.7\columnwidth]{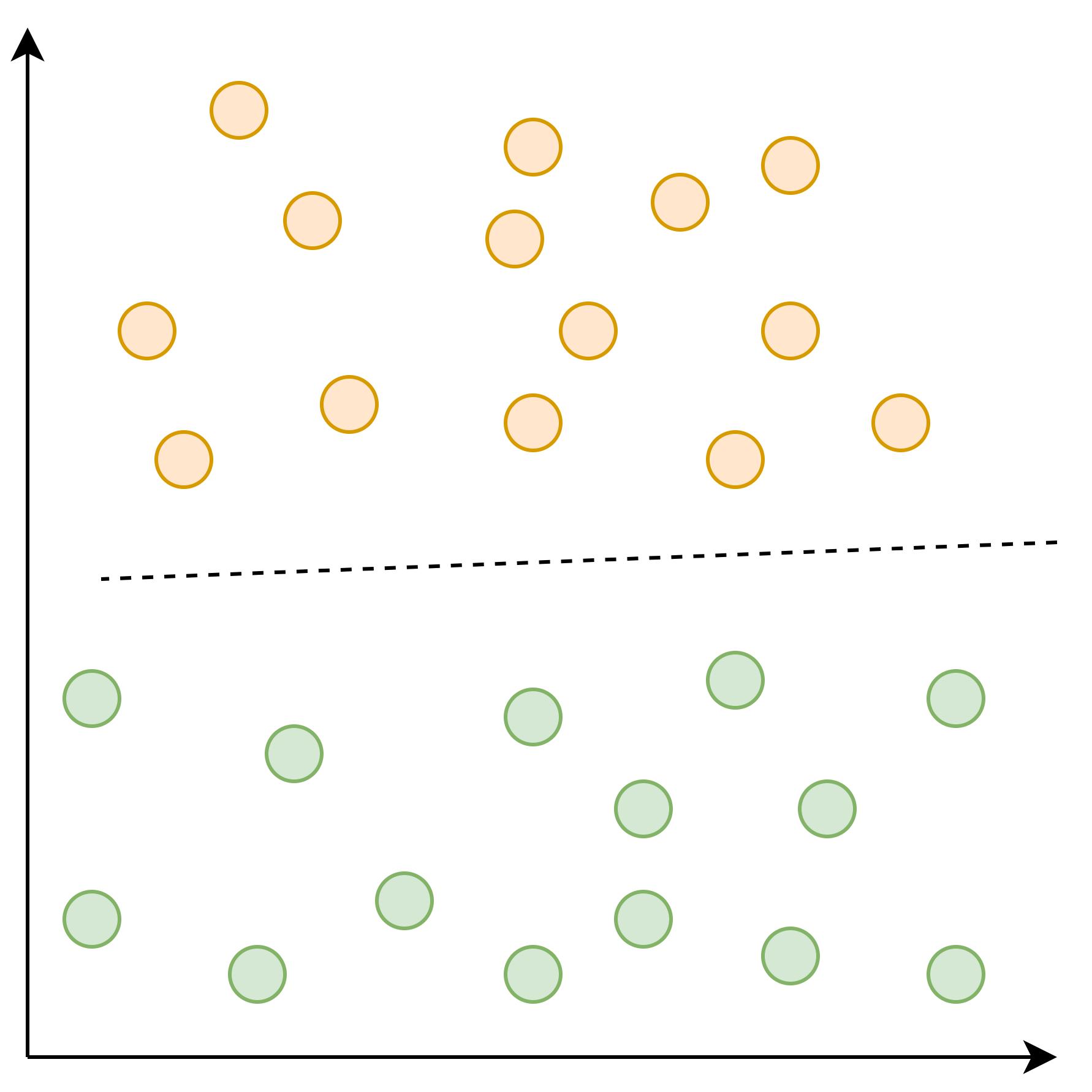}
  \caption{Real drift}
  \label{fig:sub2}
\end{subfigure} \hspace{50mm}
\begin{subfigure}{.7\columnwidth}
  \centering
  \includegraphics[width=.7\columnwidth]{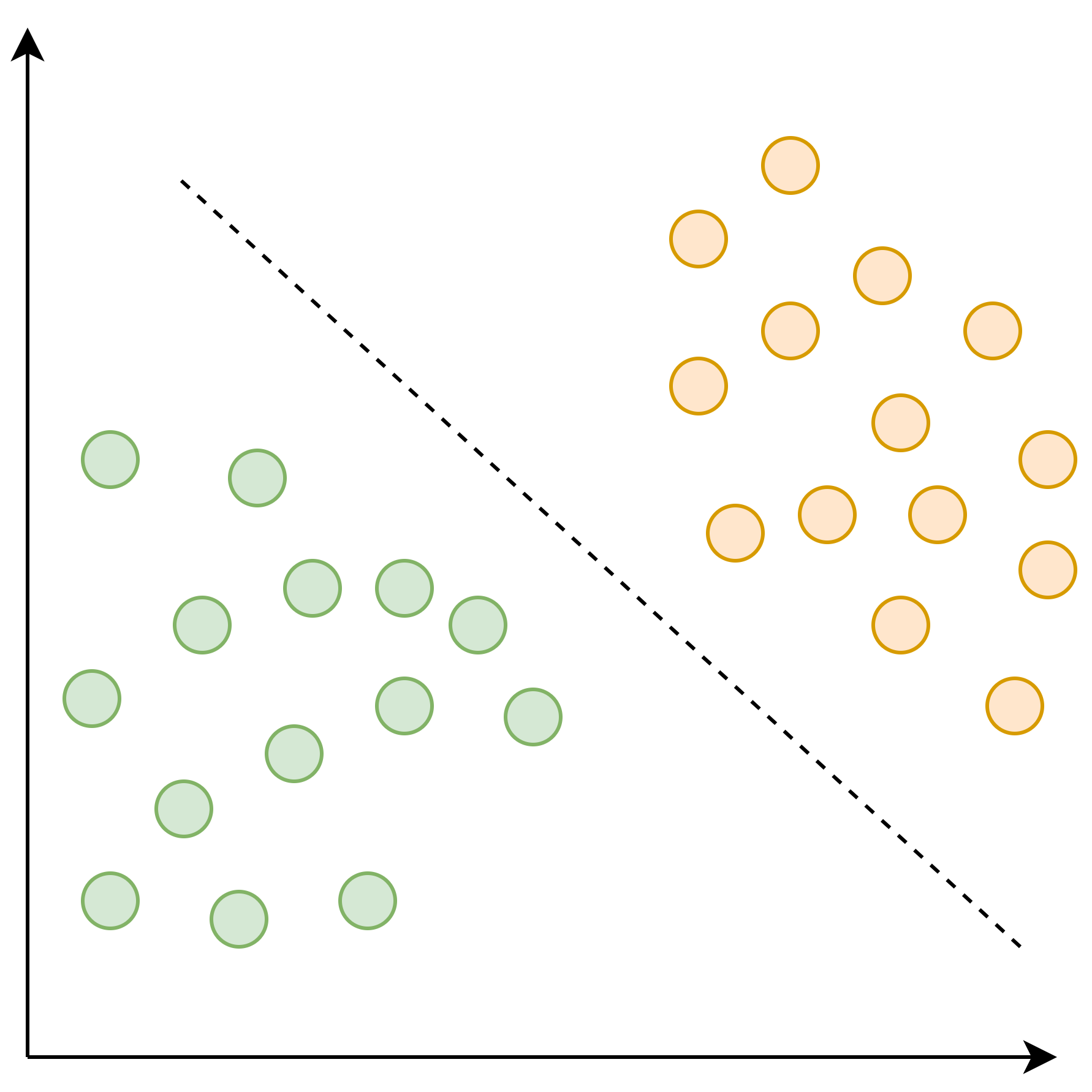}
  \caption{Virtual drift}
  \label{fig:sub1}
\end{subfigure}%
\begin{subfigure}{.7\columnwidth}
  \centering
  \includegraphics[width=.7\columnwidth]{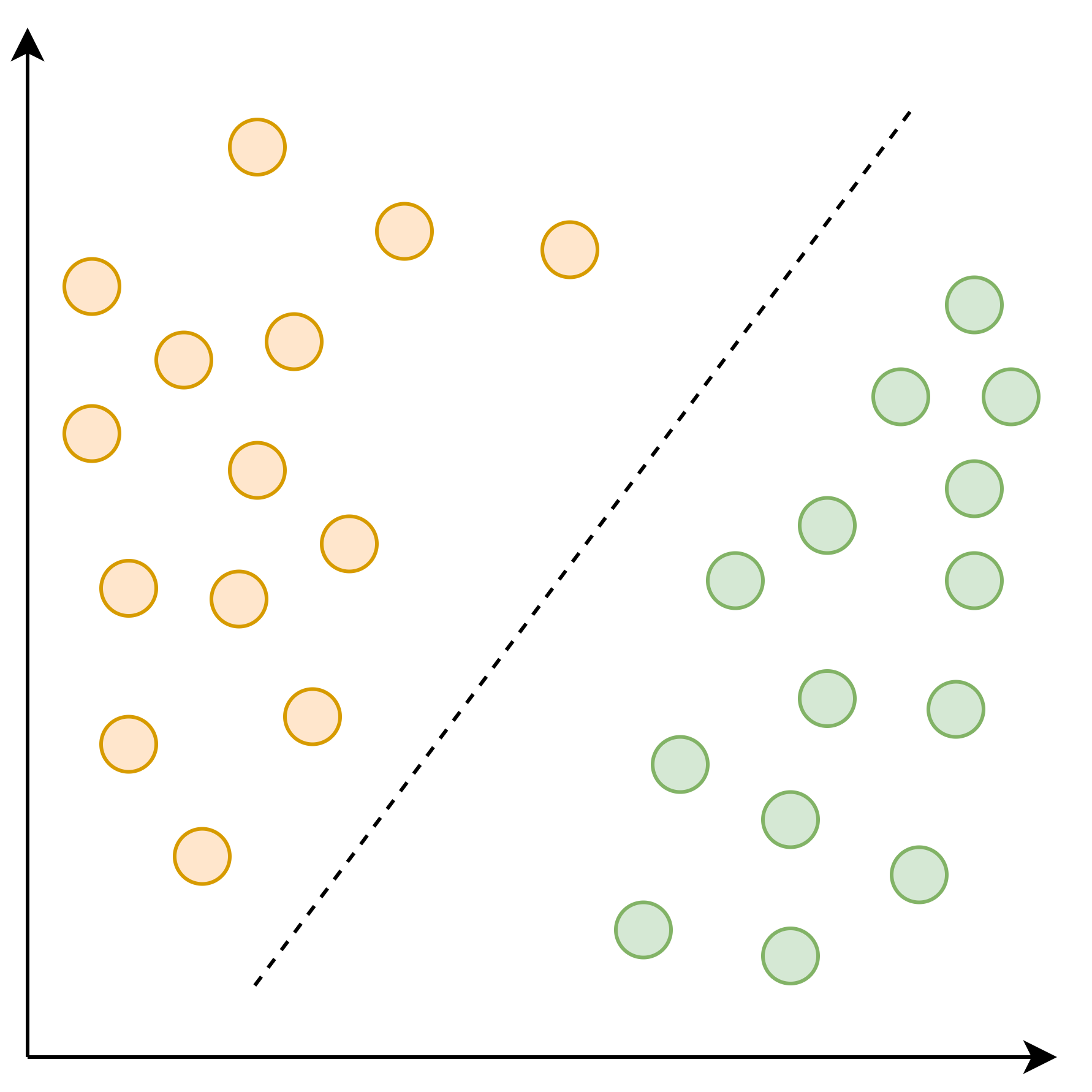}
  \caption{Rigorous drift}
  \label{fig:sub1}
\end{subfigure}
\caption{Sources of concept drift based on how they occur probabilistically. Different colors represent different classes and the dotted line is the decision boundary.}
\label{fig:drift-source}
\end{figure*}

In this study, all algorithms are used with the default parameters provided by their respective publications. In the experiments, the baseline classifier starts a new prediction model when a concept drift is detected. We should also note that, since we use a supervised classifier in our experiments, LD3 indirectly uses the ground truth labels of the data stream as it processes the predicted labels of the classifier. The reason behind this is the lack of online unsupervised multi-label classifiers. Although there are previously developed unsupervised multi-label classifiers, they are not designed for online learning, such as Wang and Zhang’s work \citep{Wang_2020_CVPR}. Therefore, we chose to use a supervised classifier.
\par
In the following sections, we first describe the problem domain, the aim of the study, and introduce the previous work in Sections 2 and 3. Then, we propose our solution in Section 4. Lastly, in Sections 5 and 6, we discuss the evaluation methodology and present our results and discussions. Section 7 concludes the paper.
\section{Problem Domain and Aim of the Study}\label{sec2}

Traditional multi-label data stream classifiers learn by performing the interleaved test-then-train method \citep{buyukccakir2018novel}, i.e, prequential training, on a stream of incoming data. However, if there is a change in data distribution, a significant decrease in predictive performance is experienced since the classifier still uses the previously learned distribution in its predictions. A concept drift detector detects the change in data distribution and alerts the classifier so that it can start working on adaptation strategies, which increases the robustness of the classifier.

Concept drift is the change in the data distribution that occurs over time. Given a data stream $S_{0,t}={d_0,...,d_t}$, in a time window $[0, t]$, where $d_i=(X_i,y_i)$ with $X_i$ being the features and $y_i$ being the labels of the \emph{i-th} data instance, concept drift is \citep{gama2014survey}:

\begin{equation}
\label{eqn:eq1}
\exists t: P_{t}(X,y) \neq P_{t+1}(X,y)
\end{equation}

According to this definition, \cite{lu2018learning} describe three potential sources of concept drift:
\begin{itemize}
    \item The change may happen due to a change in posterior probabilities $P_{t}(y \vert X)$, meaning, $P_{t}(y \vert X) \neq P_{t+1}(y \vert X)$. This is called real or actual drift and it usually results in a shift in the decision boundary, causing a significant decrease in effectiveness.
    \item If the cause of the change is $P_{t}(X) \neq P_{t+1}(X)$, while $P_{t}(y \vert X) = P_{t+1}(y \vert X)$, it is called a virtual drift because it does not cause a shift in the decision boundary.
    \item The final source is when the change occurs as both $P_t(y \vert X)$ and $P_t(X)$ change over time which is called rigorous drift \citep{gozuaccik2021concept}. 
\end{itemize}

Figure \ref{fig:drift-source} illustrates the three sources of concept drift along with the original distribution.

\begin{figure*}[!ht]
\centering
\begin{subfigure}{.6\columnwidth}
  \centering
  \includegraphics[width=.7\linewidth]{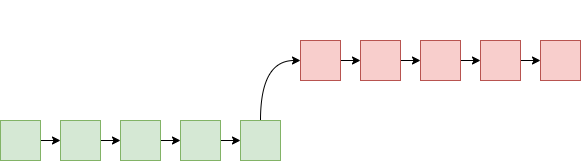}
  \caption{Sudden drift}
  \label{fig:sub1}
\end{subfigure}%
\begin{subfigure}{.6\columnwidth}
  \centering
  \includegraphics[width=.7\linewidth]{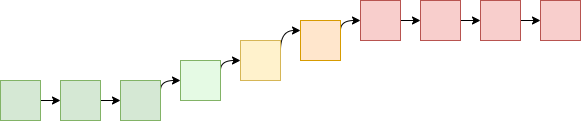}
  \caption{Incremental drift}
  \label{fig:sub2}
\end{subfigure} \hspace{50mm}
\begin{subfigure}{.6\columnwidth}
  \centering
  \includegraphics[width=.7\linewidth]{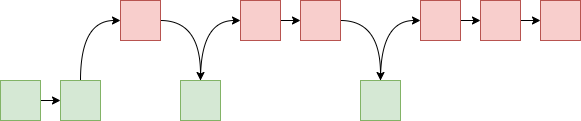}
  \caption{Gradual drift}
  \label{fig:sub1}
\end{subfigure}%
\begin{subfigure}{.6\columnwidth}
  \centering
  \includegraphics[width=.7\linewidth]{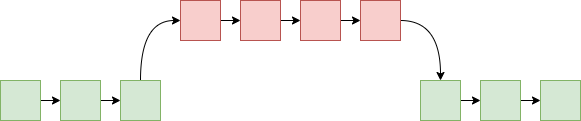}
  \caption{Reoccurring concepts}
  \label{fig:sub1}
\end{subfigure}
\begin{subfigure}{.6\columnwidth}
  \centering
  \includegraphics[width=.7\linewidth]{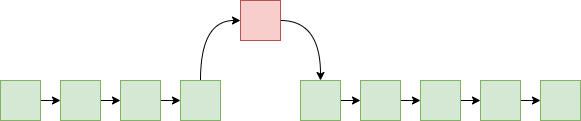}
  \caption{Outlier}
  \label{fig:sub1}
\end{subfigure}%
\caption{Types of concept drift based on what happens to the concept over time (outlier is not a drift). Different colors represent different concepts.}
\label{fig:drift-type}
\end{figure*}

\cite{lu2018learning} further define four types of concept drift in terms of how they happen over time, which are illustrated in Figure \ref{fig:drift-type}. In the figure, the vertical axis represents data distribution and the horizontal axis represents time. Depending on the nature of data, this change may be sudden, incremental, gradual, or reoccurring. For example, with the changing of seasons, climate data may show reoccurring concepts or big events may cause sudden drift in news data. 
\par
Among the types presented in Figure \ref{fig:drift-type}, outlier is not actually a drift type. Outliers usually happen as a result of noise in the data. As they are false positives, a drift detector should also account for their presence to maintain accurate detection. 
\par
In this study, we aim to design an unsupervised concept drift detector and measure its boosting impact in classifier prediction accuracy. Our method LD3 exploits dependencies among predicted labels and grants drift detection capabilities to multi-label stream classifiers with no drift detection facility. In this work, we consider a supervised classification environment. The use of LD3 with an unsupervised multi-label \citep{Wang_2020_CVPR} or a self-tuning classifier \citep{roseberry2018multi} is beyond the scope of this paper.


\begin{table*}[tp]
\caption{Error rate-based drift detection algorithm characteristics (They are all our baselines; for those with more than one version, the number of variants is given in parentheses after the detector name). The symbols $w_{hist}$ and $w_{new}$ indicate the windows for old and new data in respective order. Landmark uses the entire set of data before drift and flushes this data when drift is detected to collect the new distribution.}
\resizebox{\textwidth}{!}{\begin{tabular}{llllll}
\hline
\textBF{Category} &\textBF{Detector} & \textBF{Data Retrieval}       & \textBF{Test Statistics} & \textBF{Hypothesis Test} \\ \hline
Error rate-based        &ADWIN \cite{bifet2007learning}            & Auto cut $w_{hist}$, $w_{new}$                  &                            Error rate difference    & Hoeffding's Bound        \\                 
                        &DDM \cite{gama2004learning}               & Landmark                                  & Online Error Rate        & Distribution Estimation  \\
                        &EDDM \cite{baena2006early}              & Landmark                                  & Online Error Rate        & Distribution Estimation  \\
                        &FHDDM \cite{pesaranghader2016fast}              & Sliding $w_{hist}$, $w_{new}$                                  & Correct prediction probability        & Hoeffding's Bound  \\ 
                        &FHDDMS (2) \cite{pesaranghader2018reservoir}             & Stacked windows                                  & Correct prediction probability        & Hoeffding's Bound  \\
                        &HDDM (2) \cite{frias2014online}              & Landmark                                  & Online Error Rate        & Hoeffding's Bound  \\ 
                        &KSWIN \cite{raab2020reactive}             & Sliding $w_{hist}$, $w_{new}$                                 & Distribution distance        & KS-Test \\
                        &MDDM (3) \cite{pesaranghader2018mcdiarmid}              & Sliding $w_{hist}$, $w_{new}$                                  & Weighted mean difference        & McDiarmid's Bound  \\ 
                        &RDDM \cite{barros2017rddm}              & Landmark                                  & Online Error Rate        & Distribution Estimation  \\
                        &SeqDrift2 \cite{pears2014detecting}             & Sliding $w_{hist}$, $w_{new}$                                  & Online error rate        & Bernstein Bound  \\ \hline
Data distribution-based & LD3 (Our method)              & Sliding $w_{hist}$, $w_{new}$ &  Rank correlation         & Distribution Estimation                         \\ \hline
\end{tabular}}
\label{tab:detector-table}
\end{table*}

\section{Related Works}\label{sec3}

In the following subsections, we first briefly introduce previously developed supervised detection algorithms based on online error rate monitoring and multi-label specific algorithms. Then, previous work on some of the unsupervised concept drift detection algorithms is presented.
\subsection{Supervised Concept Drift Detection Algorithms}
Concept drift is a prevalent problem in data stream mining, where many detectors are proposed to combat it within the literature. Based on previously developed algorithms, \cite{lu2018learning} propose an overall framework for concept drift detection which is comprised of four stages: Data retrieval, data modeling, test statistics calculation, and the hypothesis test. They further divide the detectors into three categories: Error rate-based, data distribution-based, and the multiple hypothesis test. Among these three, we focus on the first category as we were unable to find available source codes for multiple hypothesis test algorithms and data distribution-based algorithms. Table \ref{tab:detector-table} displays the categories and methods used for each baseline algorithm we tested. The data modeling stage is not included in the table since all of the detectors except LD3 share the same data modeling scheme which is learner-based modeling.

\subsubsection{Error Rate-based Algorithms}
The detection algorithms in this category usually focus on the changes in the online error rate of the base classifiers, i.e, whether changes in the error rate of a classifier are statistically significant enough, the detector concludes that there is drift. Since these detectors monitor the online error rate and related metrics, they usually employ learners to model the data, which is the case for all of the algorithms presented in Section 3.1.1, and they tend to be efficient algorithms due to the simplicity of their input data (error-rate).
\par 
One of the most frequently used algorithms is the Drift Detection Method (DDM) proposed by \cite{gama2004learning}. It uses a separate classifier to check whether the change in the online error rate is significant enough, which is done through distribution estimation. If the change is statistically significant, drift is detected. There are also a large number of algorithms that build upon the DDM algorithm, which usually change the test statistics and hypothesis test stages. For instance, Early Drift Detection Method (EDDM) \citep{baena2006early} improves DDM by also considering the sample-wise distance between erroneous classifications. 
\par
In order to improve the hypothesis test stage, \cite{frias2014online} propose HDDM, which utilizes Hoeffding's inequality \citep{hoeffding1963} to obtain probabilistic guarantees to further increase effectiveness. It has two variants, namely HDDM\_A and HDDM\_W, where, in the former, they use bounded moving averages (A-test) and in the latter, they use bounding weighted moving averages (W-test). Furthermore, \cite{pesaranghader2016fast} introduce Fast Hoeffding Drift Detection Method (FHDDM) which requires the base detectors to either stay at a steady level or improve in accuracy. FHDDM checks this by monitoring the most recent probability of correct predictions instead of the error rate. \cite{pesaranghader2018reservoir} further improve FHDDM as FHDDMS by adopting a stacking window scheme of various sizes. The stacking windows are a short and long window that has overlapping content that is used to detect sudden and gradual drifts separately. It also has another variant, FHDDMS\_add, where the authors employ additive summaries for better efficiency in memory and execution time.
\par
Apart from Hoeffding's inequality-based improvements to DDM, \cite{pesaranghader2018mcdiarmid} developed McDiarmid Drift Detection Method (MDDM). MDDM is similar to HDDM as they also make improvements in hypothesis testing, by including the McDiarmid inequality to check the statistically significant differences. MDDM also implements a weighting scheme within its window where the most recent elements are given more importance. According to alternate weighting, it has three variants: MDDM\_A (Arithmetic weighting), MDDM\_E (Euler weighting), and MDDM\_G (Geometric weighting). 
\par 
Instead of modifying a stage, \cite{barros2017rddm} propose Reactive Drift Detection Method (RDDM), in which they define a type of soft concept drift based on the number of samples accumulated in the window, called RDDM drift. RDDM re-calculates their DDM-based statistics to solve problems that arise from the high number of accumulated samples in the window.
\par
Aside from DDM-based detectors, \cite{bifet2007learning} introduce ADWIN, which monitors the difference in error rate between two adaptive windows. The difference is bounded by Hoeffding's inequality and if the difference exceeds the Hoeffding bound, a drift is detected.
\par
Furthermore, \cite{pears2014detecting} propose SeqDrift2. It employs an adapting sampling strategy to sample data from a window to get old and new concepts. Then, they detect the drift by comparing the differences according to the Bernstein bound.
\par
Finally, \cite{raab2020reactive} introduce KSWIN, which detects concept drift by applying ``the Kolmogorov-Smirnov test" (KS-Test) which finds the distance between the estimated data distribution and the empirical distribution. These two distributions in KSWIN's case would be the error rates stored in a sliced sliding window for new and old data. If the distance between the distributions exceeds the confidence interval, a drift is detected.

\subsubsection{Algorithms for Multi-label Concept Drift Detection}
In the multi-label classification problem domain, concept drift detection is a very thinly researched area. To the best of our knowledge, there are two previously developed concept drift detectors. However, in both cases the authors did not provide source codes thus we were unable to compare our results with them which is the reason why we did not include them in Table \ref{tab:detector-table}. 

\par
\cite{shi2014drift} propose a method in which they use label grouping and class entropy to detect concept drift. Initially they group the labels using clustering methods. Then for each label group, they calculate the multi-label entropy values within two sliding windows where they apply a threshold method to detect concept drift.
\par 
\cite{wang2020concept} introduce DDM-FP-M, a multi-label focused concept drift detector aimed at data streams on the Internet of Things problem domain. They propose modifications to DDM in which they add false positive classifications to make it more suitable for multi-labeled data stream environments.

\subsection{Unsupervised Concept Drift Detection}
Although many concept drift detectors exist, one area that is insufficiently researched is unsupervised concept drift detection. \cite{iwashita2018overview} found that unsupervised drift detectors only make up 3\% of developed concept drift detectors. However, as Gemaque et al. point out, many of the real-world problems are better suited for unsupervised drift detection, since the swift acquisition of labels is often not possible \citep{gemaque2020overview}. Since concept drift detection requires fast detection to enable rapid recovery after drift, the topic of unsupervised concept drift detection requires more attention. 
\par 
Some of the previous works on unsupervised concept drift detection include detectors such as IKS-bdd \citep{dos2016fast}, CD-TDS \citep{koh2016cd}, Plover \citep{de2019learning}, DSDD \citep{pinage2020drift}, D3 \citep{gozuaccik2019unsupervised}, and OCDD \citep{gozuaccik2021concept}.
\par
IKS-bdd applies the Incremental Kolmogorov-Smirnov test, which is a modified Kolmogorov-Smirnov test that is better suited for online learning, to each of the data features in order to detect drift. CD-TDS detects two types of drift: Local drift and global drift. For local detection, the sample means of the new and old data are compared, bounded by the Hoeffding Bound. In the case of global drift, a pairwise statistical test is applied to find differences between two tree structures representing new and old data.
\par 
Plover monitors the input data behavior  and detects changes based on observed instabilities. Moreover, DSDD detects drifts based on classification error simulation using an ensemble of classifiers.
\par
Furthermore, Gözüaçık et al. introduce D3, a drift detector that utilizes a discriminative classifier that is trained on auto labeled samples based on how recent a sample was seen within a window (``0" for old and ``1" for new samples). Drift detection is made by measuring the AUC score of the classifier. Likewise, OCDD uses a one-class classifier to distinguish between changing concepts, in which drift is detected when the ratio of false predictions is higher than a threshold.
\par
Our work proposes an unsupervised drift detection algorithm that utilizes the predicted labels of the paired classifier; however, we were unable to compare our detector with the described unsupervised algorithms, excluding D3 and OCDD, as we could not find readily available source code. 
\par
Furthermore, D3 and OCDD are not compared with LD3 since we only used original codes provided by the frameworks we used. The provided source codes are incompatible with multi-label classification and we chose not to modify the original code for this reason.

\section{Our Method: Label Dependency Drift Detector}\label{sec4}

In this study, we propose LD3, a data distribution-based, unsupervised concept drift detector which exploits the dependencies among predicted labels. It works alongside any online multi-label classifier that does not have inherent drift detection properties to assist in handling the changes in data distribution. We evaluate the changes in the label dependencies through the predicted labels provided by the paired model, and if a significant change occurs in the dependencies, we detect concept drift. In the remainder of this section, Algorithm 1 is referenced for explanations.

\begin{table}[!h]
\centering
\caption{Symbols table for Algorithm 1.}
\begin{tabular}{ll}
\hline
\textBF{Symbol} & \textBF{Description}  \\ \hline
$l$               & \multicolumn{1}{l}{Predicted label} \\
$L$               & \multicolumn{1}{l}{Threshold for number of anomalies} \\
$t$               & \multicolumn{1}{l}{Standard deviation multiplier} \\
$w$               & \multicolumn{1}{l}{Number of samples in a window} \\ 
$W_{new}$               & \multicolumn{1}{l}{Label window for new samples} \\ 
$W_{old}$               & \multicolumn{1}{l}{Label window for old samples} \\ 
$W_{corr}$               & \multicolumn{1}{l}{Past correlation window}
\\ \hline
\end{tabular}
\end{table}

\begin{algorithm*} []
\caption{LD3: Label Dependency Drift Detector}
\begin{algorithmic}
\State Initialize \boldmath $W_{new}$, $W_{old}$ and $W_{corr}$ as $\emptyset$ 
\Procedure{LD3}{$l$, $t$, $w$, $L$} 
    \State \boldmath{$drift$} $\leftarrow$ False 
    \State $l^{\prime} \leftarrow $ Insert\_Element($l$, $W_{new}$) $\quad \cdot \quad \cdot \quad \cdot \quad \cdot \quad \cdot \quad \cdot \quad \cdot \quad \cdot \quad \cdot \quad \cdot \quad \cdot \quad \cdot \quad \cdot \quad \cdot \quad$ \Comment{$l^{\prime}$ is the last element removed}
    \If{$\vert W_{new} \vert$ \unboldmath = \boldmath $w$}
        \State Insert\_Element($l^{\prime}$, $W_{old}$)
    \EndIf
    \If{$\vert W_{new} \vert$ \unboldmath $\neq$ \boldmath $w$ and $\vert W_{old} \vert$ \unboldmath $\neq$ \boldmath $w$} 
        \State return $drift$
    \Else
        \State $M_{new}$ $\leftarrow$ Co-occurrence matrix created from $W_{new}$ 
        \State $M_{old}$ $\leftarrow$ Co-occurrence matrix created from $W_{old}$ 
        \State $R_{new}$ $\leftarrow$ Reciprocal ranking of $M_{new}$ $\quad \cdot \quad \cdot \quad \cdot \quad \cdot \quad \cdot \quad \cdot \quad \cdot \quad \cdot \quad \cdot \quad \cdot \quad \cdot \quad \cdot \quad$ \Comment{Global ranking for $W_{new}$}
        \State $R_{old}$ $\leftarrow$ Reciprocal ranking of $M_{old}$ $\quad \cdot \quad \cdot \quad \cdot \quad \cdot \quad \cdot \quad \cdot \quad \cdot \quad \cdot \quad \cdot \quad \cdot \quad \cdot \quad \cdot \quad \cdot$ \Comment{Global ranking for $W_{old}$}
        \State $C$   $\leftarrow$  WS($R_{new}$, $R_{old}$) $\quad \cdot \quad \cdot \quad \cdot \quad \cdot \quad \cdot \quad \cdot \quad \cdot \quad \cdot \quad \cdot \quad \cdot \quad \cdot \quad \cdot \quad \cdot$ \Comment{Rank Correlation of $R_{new}$ and $R_{old}$}
        \State Insert\_Element($C$, $W_{corr}$) 
        \If{$\vert W_{corr} \vert$ \unboldmath $\neq$ \boldmath $w$}
            \State return $drift$
        \EndIf
        \State $len \leftarrow$ Sigma\_Rule($W_{corr}$, $t$) $\quad \cdot \quad \cdot \quad \cdot \quad \cdot \quad \cdot \quad \cdot \quad \cdot \quad \cdot \quad \cdot \quad \cdot \quad \cdot \quad \cdot \quad \cdot$ \Comment{Number of anomalies in $W_{corr}$}
        \If{$len > L$} 
            \State $drift \leftarrow$ True 
            \State Clear windows $W_{new}, W_{old}, W_{corr}$
        \EndIf
    \EndIf
    \State return $drift$
\EndProcedure
\end{algorithmic}
\end{algorithm*}

\subsection{Evaluating the Changes in the Label Dependencies}
In an ideal environment, labels can be chosen such that each label is independently distributed, i.e, dependencies do not exist. However, in real-world datasets, this is usually not the case, as some labels tend to occur more frequently together (e.g, in movie categories, comedy and drama tags occur more commonly together than comedy and thriller tags). In our study, we hypothesize that this correlation causes dependencies, and changes in these dependencies are a precursor to concept drift.
\par
Given a multi-label classifier, we buffer the labels predicted by the classifier in two fixed-sized moving windows, one each for the new and old data. Apart from providing up-to-date statistics about label distribution, these windows allow the classifier to boot itself up, i.e, learn enough of the data distribution to generate consistent predictions, as the windows are filled (Alg. 1, lines 4-6), which functions as a warmup scheme.

\par

After the windows are full, we first generate two co-occurrence matrices from the windows (Alg. 1, lines 10-11). The matrices are obtained by counting the number of times each class label occurs as ``1" alongside other labels. The generated matrices are then ranked within each row, which we call local ranking, by creating a ranking for each label based on their co-occurrence frequencies.

\par

Following the local ranking, the ranks are aggregated by utilizing a data fusion algorithm to obtain a representation of label dependencies for new and old samples (Alg. 1, lines 12-13). We call the resulting aggregated ranking as global ranking. Through this, we obtain the most influential labels among all of them and use these labels to monitor changes in the stream. We use reciprocal rank fusion, which is seen in Equation \ref{eqn:reciprank}, where $r_i$ is the global ranking of a class label within the predicted label $l$, which is denoted by $l_i$. Moreover, we use $n$ to represent the number of classes and $r_{ij}$ for the local ranking of $l_i$, where $\{j \,\vert \, 1 \leq j \leq n\}$. We justify our selection of reciprocal rank fusion, rather than some other commonly used approaches, by experiments in Section 6.3.

\begin{equation}
\label{eqn:reciprank}
r_i = \frac{1}{\sum_{j=1}^{n} \frac{1}{r_{ij}}} \qquad \{i\,\vert \, 1 \leq i \leq n\}
\end{equation}

\subsection{Measuring Similarity Between Two Rankings}
Subsequent to the global ranking, we calculate the rank correlation between the two global rankings we obtained (Alg. 1, line 14), which evaluates the similarity between rankings. We use the WS coefficient as our rank correlation measure which is a ranking similarity method developed by \cite{salabun2020new}. It calculates the weighted similarity between two rankings and returns a value bounded within $[-1, 1]$, where two identical rankings are scored as 1, whereas the score is -1 for the opposite case, from which can be deduced Equation \ref{eqn:wscoeff}. In this equation, $R_{xi}$ and $R_{yi}$ represent the rank position of a label $l_i$ within $R_{new}$ and $R_{old}$. 

\begin{equation}
\label{eqn:wscoeff}
C = 1 - \sum_{i=1}^{n}\left(2^{-R_{xi}}\cdot \frac{\vert R_{xi} - R_{yi}\vert }{max\{\vert 1-R_{xi}\vert,\vert n-R_{xi} \vert\}}\right) 
\end{equation}

Within the summation, $2^{-R_{xi}}$ is the weight of the label $l_i$ which ensures that higher ranking labels have more influence. The numerator $\vert R_{xi} - R_{yi}\vert$ is the ranking distance of $l_i$ within the two rankings and the denominator scales this distance.


We chose to use the WS coefficient instead of other popular rank correlation measures such as Kendall's $\tau$ \citep{kendall1938new} or Spearman's $\rho$ \citep{spearman1987proof} because it provides a way to weigh the labels. Since we measure the influences labels have on each other, a change in the higher ranked labels is more important than other labels. Furthermore, it measures the similarity based on the distance between the two rankings. In a possibly volatile environment like a data stream, rankings could change by a few places temporarily; however, if we measure the similarity based on distance, such irregularities are tolerated more easily, which is why measures such as \emph{weighted} $\tau$ \citep{vigna2015weighted} are not suitable for our case.

\begin{figure*}[tp]
    \centering
    \includegraphics[width=\textwidth]{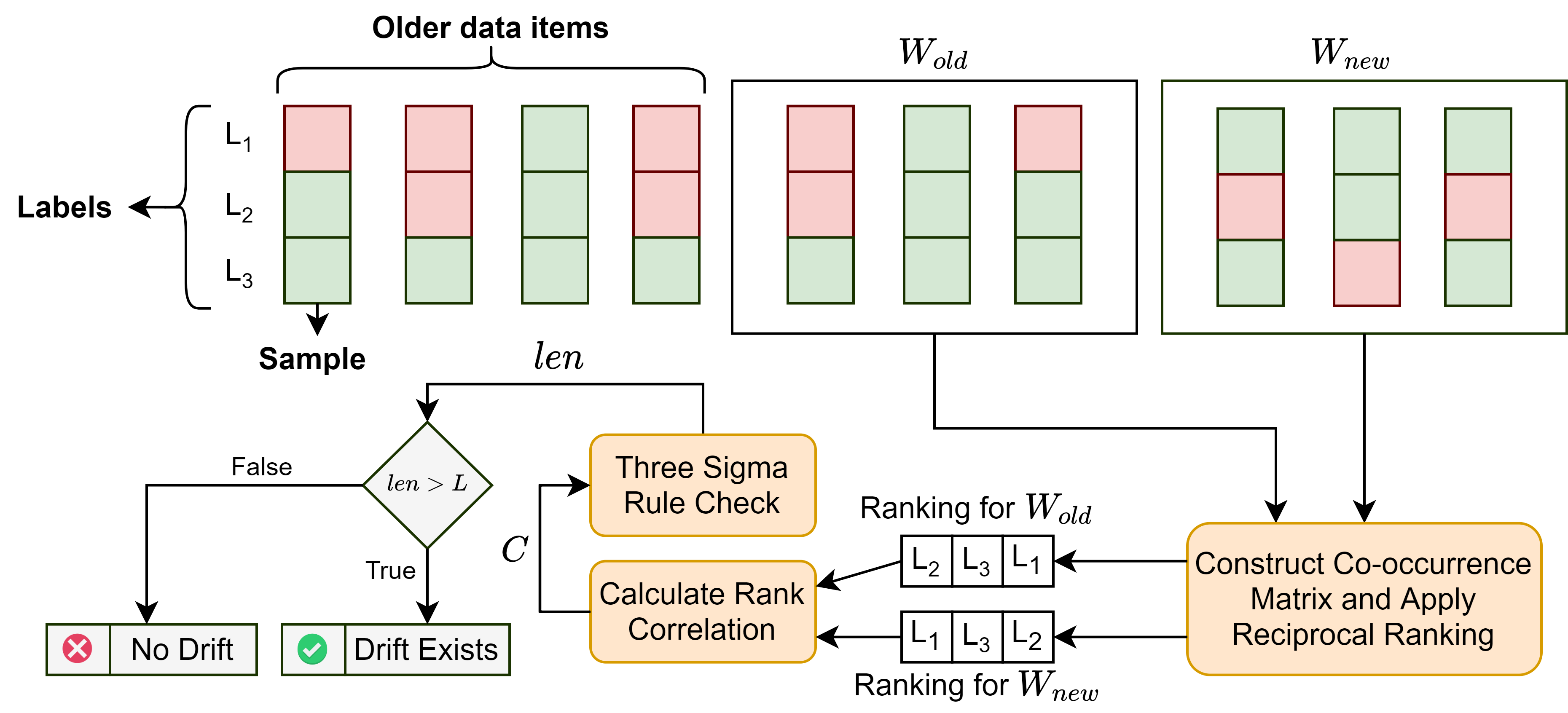}
    \caption{Workflow of LD3. Red boxes represent false labels (0) and green boxes represent true labels (1).}
    \label{fig:ld3}
\end{figure*}
\begin{figure*}[bp]
    \centering
    \includegraphics[width=0.6\textwidth]{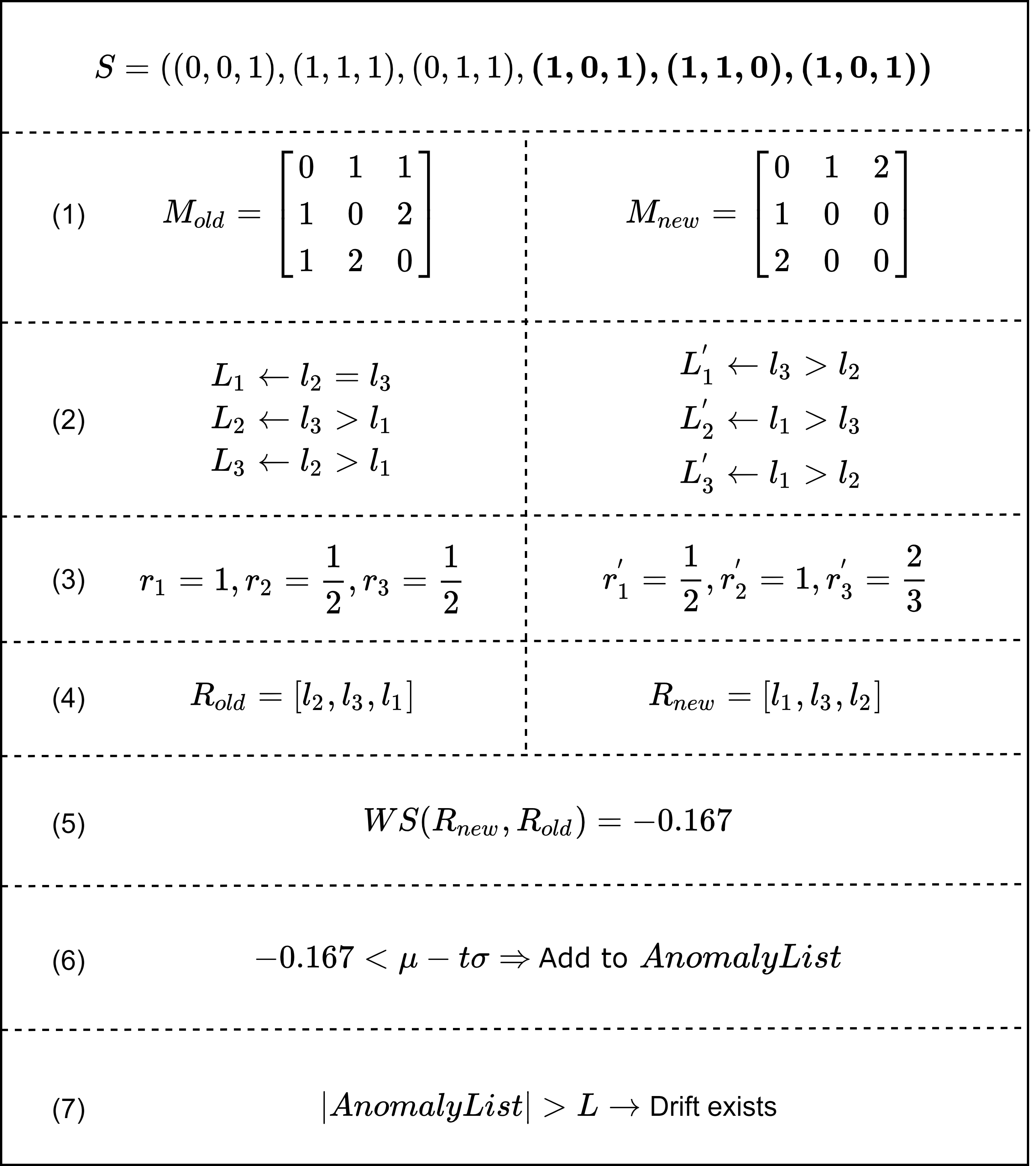}
    \caption{A simple numerical example of LD3 where $\mu$ and $\sigma$ are the mean and standard deviation of the samples within $W_{corr}$, t is the standard deviation multiplier for the three sigma rule \citep{pukelsheim1994three}, and L is the number of anomalies threshold. Recently arrived predicted labels are highlighted in bold. The calculations are explained in more detail in Appendix \ref{secA1}.}
    \label{fig:num_example}
\end{figure*}

Although the generated rank correlation value implies whether the current ranking indicates a drift, it needs to be checked for statistical significance. This is done by utilizing the three sigma rule \citep{pukelsheim1994three}, which translates to a simple left tailed test for our case (Alg. 1, line 15-19). A separate moving window ($W_{corr}$) is utilized to accumulate the past rank correlations. If the current correlation is less than the mean ($\mu$) within the window by $t$ times the standard deviation ($\sigma$), it is considered an anomaly. However, such anomalies may be the result of noise in the data, i.e, outliers described in Section 2. To prevent false detections, we store these anomalies in a list, and if the length of this list is greater than a chosen length $L$, the detector signals that a drift has been encountered. The three sigma rule is used here to increase the computational efficiency of the algorithm as the desired result is obtained without having to estimate the distribution of the recent data in its entirety.
\par

The general representation of all of the stages is illustrated in Figure \ref{fig:ld3}. In addition, a simple numerical example of LD3 is given in Figure \ref{fig:num_example}. In this figure, $M_{old}$ and $M_{new}$ are obtained by counting the co-occurrences of labels. Then, local rankings are obtained which are represented as $L_i$ and $L^{\prime}_i$ for each label within $M_{old}$ and $M_{new}$. The variables $r_i$ and $r^{\prime}_i$ are reciprocal ranking results for old and new samples for each label.

\section{Evaluation and Experimental Setup}\label{sec5}
\subsection{Datasets}
For evaluation, we used nine well-known real-world multi-label datasets, which we obtained in MEKA \citep{read2016meka} formats, and three synthetic data streams \citep{montiel2018scikit}. Table \ref{tab:datasets} represents the datasets and their properties. The datasets are chosen among the ones tested in the study of \cite{roseberry2018multi}. We used all datasets from that list shown to have drift by any of the tested detectors and had better effectiveness results than a baseline classifier without drift detection functionalities. Since we measure the detectors' benefit based on the effectiveness improvement on the classifier, if the baseline classifier shows higher predictive performance, we conclude that detectors only make false positive detections. In such a case, we assume the dataset does not have drift.

\begin{table}[h]
\begin{center}
\begin{minipage}{\columnwidth}
\caption[Table of multi-label datasets used in the experiments. The upper group contains the real-world datasets. $N$ represents the number of samples, $D$ is the number of features, $n$ is the number of labels and LC($\mathcal{D}$) and LD($\mathcal{D}$) represents label cardinality and label density which are the average number of true labels of the samples in dataset $\mathcal{D}$ and LC($\mathcal{D}$) divided by the number of labels, indicating the average label cardinality per label \citep{tsoumakas2007multi}.]{Table of multi-label datasets used in the experiments\footnote{The real-world datasets can be accessed from: http://www.uco.es/kdis/mllresources/}. The upper group contains the real-world datasets. $N$ represents the number of samples, $D$ is the number of features, $n$ is the number of labels and LC($\mathcal{D}$) and LD($\mathcal{D}$) represents label cardinality and label density which are the average number of true labels of the samples in dataset $\mathcal{D}$ and LC($\mathcal{D}$) divided by the number of labels, indicating the average label cardinality per label \citep{tsoumakas2007multi}.}\label{tab:datasets}
\resizebox{\columnwidth}{!}{\begin{tabular}{lrrrrrr}
\hline
\textBF{Dataset Name}   & \textBF{Domain}   & \boldmath{$N$}      & \boldmath{$D$}     & \boldmath{$n$}   & \textBF{LC($\mathcal{D}$)}    & \textBF{LD($\mathcal{D}$)}    \\ \hline
20NG                    & Text   & 19,300  & 1,006  & 20   & 1.029 & 0.051 \\ 
Birds          & Audio         & 645     & 260    & 19   & 1.014 & 0.053 \\
Enron             & Text    & 1,702   & 1,001  & 53   & 3.378 & 0.064 \\
EukaryotePseAAC      & Biology  & 7,766    & 440    & 22   & 1.146 & 0.052 \\
Imdb             & Text      & 120,900 & 1,001  & 28   & 2.000 & 0.071 \\
Ohsumed        & Text       & 13,930  & 1,002  & 23   & 1.663 & 0.072 \\
PlantPseAAC         & Biology   & 978     & 440    & 12   & 1.079 & 0.090 \\
Tmc2007-500    &  Text        & 28,600  & 500    & 22   & 2.220 & 0.101 \\
Yeast        & Biology          & 2,417    & 103    & 14   & 4.237 & 0.303 \\
\hline
Synthetic Sudden Drift     & Generic & 20,000  & 200   & 50  & 1.587 & 0.0397 \\
Synthetic Incremental Drift & Generic & 20,000  & 200   & 50  & 1.588 & 0.0397 \\ 
Synthetic Re-occurring Drift & Generic & 20,000  & 200   & 50  & 1.588 & 0.0397 \\\hline
\end{tabular}}
\end{minipage}
\end{center}
\end{table}

The three synthetic data streams are generated to measure the difference in effectiveness results among sudden, incremental, and reoccurring drifts, which are generated by changing between three different multi-label data streams that have different label cardinalities (for reoccurring concepts, the third data stream has the same properties and data distribution as the first data stream). The change between streams is smoothed by the sigmoid function over a specified amount of samples. To simulate sudden drift, we applied the change over one sample whereas in the reoccurring and incremental drifted streams the change is applied over 500 samples. The data streams have 20,000 samples and the drifts are at sample positions 4,000 and 10,000. The properties of the streams are shown in Table \ref{tab:datasets}.

\subsection{Experimental Setup and Evaluation Metrics}

The experiments are performed using the Scikit-Multiflow \citep{montiel2018scikit} and Tornado \citep{pesaranghader2018reservoir} frameworks. Scikit-multiflow is employed for synthetic data stream generation and it contains six (ADWIN, DDM, EDDM, HDDM\_A, HDDM\_W, KSWIN) of the tested baseline detectors. Moreover, the Tornado framework is utilized as it contains the remaining eight baseline detectors. 
\par
To demonstrate the effectiveness benefits of the detectors, a Classifier Chain (CC) \citep{read2011classifier} using Gaussian Naive Bayes (NB) classifiers \citep{john1995estimating} is used as a base classifier. CCs build binary classifiers for each class label that are linked in a chain to preserve inter-label dependencies. The reason for using a CC is that it provides accurate results with reasonably fast execution. The same reasoning applies to the use of NBs as base classifiers and it is also a popular base classifier that is frequently used in the literature \citep{pintas2021feature}. Furthermore, our concept drift adapting strategy resets the classifier if drift is detected.

\par

\cite{gama2014survey} propose three possible metrics for change detection algorithms: (1) Probability of true change detection, (2) Probability of false alarms, and (3) Delay of detection. In our experiments and discussions, in order to incorporate the suggested measures, we first perform effectiveness tests on the 12 datasets we presented. Then, through plots, we make an in-depth analysis on the obtained effectiveness measures on the top four performing detectors, in which we discuss the effects of the false detections (alarms) and true detections. Lastly, to measure the impact of detection delay, we perform experiments using six synthetic data streams with varying drift speeds, i.e, how fast does the change happen between old and new concepts. This experiment allows us to assess the detectors' response within different drift environments.

\begin{table*}[bp]
\begin{center}
\begin{minipage}{\textwidth}
\caption[Experiment results for the detectors. ND is a classifier without any detection methods for the baseline. Average result is calculated from the average result of each algorithm, except LD3, for a dataset. \emph{LD3 Imp.} represents LD3's improvement over the average. \emph{Avg. Imp.} illustrates the average improvement for each metric. The best results are highlighted in bold.]{Experiment results for the detectors\footnote[2]{We will make the implementation of LD3 available on Github after the review process.}. ND is a classifier without any detection methods for the baseline. Average result is calculated from the average result of each algorithm, except LD3, for a dataset. \emph{LD3 Imp.} represents LD3's improvement over the average. \emph{Avg. Imp.} illustrates the average improvement for each metric. The best results are highlighted in bold.}\label{tab:results}

\begin{subtable}{\textwidth}
\caption{Example-based accuracy and Hamming score results}\label{cst1}
\centering

\resizebox{\textwidth}{!}{\begin{tabular}{lccccccccccccr}
\hline
\textBF{Detector} &
  \textBF{20NG} &
  \textBF{Birds} &
  \textBF{Enron} &
  \textBF{\begin{tabular}[c]{@{}c@{}}Eukatyote\\ PseAAC\end{tabular}} &
  \textBF{Imdb} &
  \textBF{Ohsumed} &
  \textBF{\begin{tabular}[c]{@{}c@{}}Plant\\ PseAAC\end{tabular}} &
  \textBF{\begin{tabular}[c]{@{}c@{}}Tmc2007-\\ 500\end{tabular}} &
  \textBF{Yeast} &
  \textBF{\begin{tabular}[c]{@{}c@{}}Synthetic\\ Incremental\end{tabular}} &
  \textBF{\begin{tabular}[c]{@{}c@{}}Synthetic\\ Reoccurring\end{tabular}} &
  \multicolumn{1}{c|}{\textBF{\begin{tabular}[c]{@{}c@{}}Synthetic\\ Sudden\end{tabular}}} &
  \textBF{\begin{tabular}[c]{@{}r@{}}Average\\ Rank\end{tabular}} \\ \hline
\multicolumn{14}{c}{\textBF{Example-based Accuracy}} \\ \hline
LD3 &
  \textBF{0.1616} &
  \textBF{0.0992} &
  \textBF{0.1403} &
  \textBF{0.1946} &
  \textBF{0.1140} &
  \textBF{0.0693} &
  \textBF{0.3198} &
  \textBF{0.2029} &
  \textBF{0.3780} &
  \textBF{0.0788} &
  \textBF{0.0808} &
  \multicolumn{1}{c|}{\textBF{0.0788}} &
  \textBF{1.00} \\
ADWIN &
  0.1386 &
  0.0550 &
  0.1245 &
  0.0327 &
  0.1093 &
  0.0361 &
  0.1431 &
  0.1962 &
  0.3622 &
  0.0724 &
  0.0730 &
  \multicolumn{1}{c|}{0.0724} &
  8.46 \\
DDM &
  0.0525 &
  0.0550 &
  0.1255 &
  0.0327 &
  0.0799 &
  0.0364 &
  0.1126 &
  0.1529 &
  0.3557 &
  0.0749 &
  0.0759 &
  \multicolumn{1}{c|}{0.0749} &
  10.58 \\
EDDM &
  0.0654 &
  0.0550 &
  0.1355 &
  0.0327 &
  0.0894 &
  0.0368 &
  0.1272 &
  0.1593 &
  0.3707 &
  0.0787 &
  0.0762 &
  \multicolumn{1}{c|}{0.0744} &
  6.25 \\
FHDDM &
  0.1453 &
  0.0550 &
  0.1236 &
  0.0327 &
  0.0797 &
  0.0381 &
  0.1240 &
  0.1555 &
  0.3622 &
  0.0659 &
  0.0761 &
  \multicolumn{1}{c|}{0.0659} &
  9.83 \\
FHDDMS &
  0.1444 &
  0.0550 &
  0.1236 &
  0.0327 &
  0.0797 &
  0.0381 &
  0.1136 &
  0.1555 &
  0.3622 &
  0.0659 &
  0.0761 &
  \multicolumn{1}{c|}{0.0659} &
  10.42 \\
FHDDMS\_Add &
  0.1483 &
  0.0550 &
  0.1239 &
  0.0327 &
  0.0797 &
  0.0373 &
  0.1165 &
  0.1555 &
  0.3622 &
  0.0659 &
  0.0761 &
  \multicolumn{1}{c|}{0.0659} &
  90.956 \\
HDDM\_A &
  0.1505 &
  0.0550 &
  0.1371 &
  0.0327 &
  0.0797 &
  0.0367 &
  0.1235 &
  0.1555 &
  0.3622 &
  0.0698 &
  0.0761 &
  \multicolumn{1}{c|}{0.0722} &
  8.08 \\
HDDM\_W &
  0.1469 &
  0.0550 &
  0.1371 &
  0.0327 &
  0.0797 &
  0.0367 &
  0.1070 &
  0.1555 &
  0.3622 &
  0.0659 &
  0.0761 &
  \multicolumn{1}{c|}{0.0659} &
  9.79 \\
KSWIN &
  0.1438 &
  0.0550 &
  0.1245 &
  0.0327 &
  0.0797 &
  0.0590 &
  0.1793 &
  0.1555 &
  0.3622 &
  0.0659 &
  0.0761 &
  \multicolumn{1}{c|}{0.0659} &
  8.67 \\
MDDM\_A &
  0.1396 &
  0.0550 &
  0.1237 &
  0.0327 &
  0.0797 &
  0.0381 &
  0.1218 &
  0.1555 &
  0.3622 &
  0.0659 &
  0.0761 &
  \multicolumn{1}{c|}{0.0659} &
  10.25 \\
MDDM\_E &
  0.1419 &
  0.0550 &
  0.1242 &
  0.0327 &
  0.0797 &
  0.0381 &
  0.1180 &
  0.1555 &
  0.3622 &
  0.0659 &
  0.0761 &
  \multicolumn{1}{c|}{0.0659} &
  10.00 \\
MDDM\_G &
  0.1419 &
  0.0550 &
  0.1242 &
  0.0327 &
  0.0797 &
  0.0381 &
  0.1180 &
  0.1555 &
  0.3622 &
  0.0659 &
  0.0761 &
  \multicolumn{1}{c|}{0.0659} &
  10.00 \\
RDDM &
  0.1562 &
  0.0574 &
  0.1237 &
  0.1225 &
  0.1130 &
  0.0382 &
  0.2225 &
  0.2016 &
  0.3622 &
  0.0722 &
  0.0746 &
  \multicolumn{1}{c|}{0.0720} &
  5.38 \\
SEQDRIFT2 &
  0.1190 &
  0.0550 &
  0.1283 &
  0.0327 &
  0.0797 &
  0.0363 &
  0.1709 &
  0.1555 &
  0.3622 &
  0.0717 &
  0.0739 &
  \multicolumn{1}{c|}{0.0750} &
  9.25 \\
ND &
  0.1469 &
  0.0550 &
  0.1371 &
  0.0327 &
  0.0797 &
  0.0608 &
  0.1369 &
  0.1555 &
  0.3622 &
  0.0659 &
  0.0761 &
  \multicolumn{1}{c|}{0.0659} &
  8.08 \\ \hline
Average Result &
  0.1321 &
  0.0552 &
  0.1278 &
  0.0387 &
  0.0846 &
  0.0403 &
  0.1357 &
  0.1614 &
  0.3623 &
  0.0689 &
  0.0756 &
  \multicolumn{1}{c|}{0.0689} &
  \textBF{Avg. Imp.} \\ \hline
LD3 Imp. (\%) &
  22.3 &
  79.7 &
  9.8 &
  402.8 &
  34.8 &
  72.0 &
  135.7 &
  25.7 &
  4.3 &
  14.4 &
  6.9 &
  \multicolumn{1}{c|}{14.4} &
  68.6 \\ \hline
\multicolumn{14}{c}{\textBF{Hamming Score}} \\ \hline
LD3 &
  0.8578 &
  \textBF{0.7302} &
  \textBF{0.8033} &
  \textBF{0.7979} &
  \textBF{0.7673} &
  \textBF{0.8023} &
  \textBF{0.8559} &
  \textBF{0.7340} &
  0.7193 &
  0.8098 &
  0.7670 &
  \multicolumn{1}{c|}{\textBF{0.8306}} &
  \textBF{1.50} \\
ADWIN &
  0.8426 &
  0.4996 &
  0.7428 &
  0.5773 &
  0.7499 &
  0.6520 &
  0.7765 &
  0.6867 &
  0.6970 &
  0.7396 &
  0.7342 &
  \multicolumn{1}{c|}{0.7408} &
  6.67 \\
DDM &
  0.2271 &
  0.4996 &
  0.7398 &
  0.5773 &
  0.7000 &
  0.1627 &
  0.5688 &
  0.5019 &
  0.7087 &
  \textBF{0.8283} &
  \textBF{0.8016} &
  \multicolumn{1}{c|}{0.7888} &
  10.17 \\
EDDM &
  0.5486 &
  0.4996 &
  0.7594 &
  0.5773 &
  0.7039 &
  0.4412 &
  0.5984 &
  0.6007 &
  \textBF{0.7221} &
  0.7969 &
  0.7743 &
  \multicolumn{1}{c|}{0.7830} &
  8.00 \\
FHDDM &
  0.8550 &
  0.4996 &
  0.7423 &
  0.5773 &
  0.7088 &
  0.7520 &
  0.6887 &
  0.5069 &
  0.6970 &
  0.6773 &
  0.6971 &
  \multicolumn{1}{c|}{0.6781} &
  9.21 \\
FHDDMS &
  0.8599 &
  0.4996 &
  0.7423 &
  0.5773 &
  0.7088 &
  0.7485 &
  0.6463 &
  0.5069 &
  0.6970 &
  0.6773 &
  0.6971 &
  \multicolumn{1}{c|}{0.6781} &
  9.67 \\
FHDDMS\_Add &
  0.8340 &
  0.4996 &
  0.7404 &
  0.5773 &
  0.7088 &
  0.7490 &
  0.6575 &
  0.5069 &
  0.6970 &
  0.6773 &
  0.6971 &
  \multicolumn{1}{c|}{0.6781} &
  10.54 \\
HDDM\_A &
  0.8069 &
  0.4996 &
  0.7632 &
  0.5773 &
  0.7088 &
  0.4764 &
  0.6022 &
  0.5069 &
  0.6970 &
  0.7026 &
  0.6971 &
  \multicolumn{1}{c|}{0.7412} &
  9.38 \\
HDDM\_W &
  0.7953 &
  0.4996 &
  0.7632 &
  0.5773 &
  0.7088 &
  0.5013 &
  0.5768 &
  0.5069 &
  0.6970 &
  0.6773 &
  0.6971 &
  \multicolumn{1}{c|}{0.6781} &
  10.58 \\
KSWIN &
  0.8343 &
  0.4996 &
  0.7445 &
  0.5773 &
  0.7088 &
  0.6916 &
  0.7489 &
  0.5069 &
  0.6970 &
  0.6773 &
  0.6971 &
  \multicolumn{1}{c|}{0.6781} &
  9.46 \\
MDDM\_A &
  0.8518 &
  0.4996 &
  0.7415 &
  0.5773 &
  0.7088 &
  0.7520 &
  0.6835 &
  0.5069 &
  0.6970 &
  0.6773 &
  0.6971 &
  \multicolumn{1}{c|}{0.6781} &
  9.67 \\
MDDM\_E &
  0.8504 &
  0.4996 &
  0.7428 &
  0.5773 &
  0.7088 &
  0.7523 &
  0.6714 &
  0.5069 &
  0.6970 &
  0.6773 &
  0.6971 &
  \multicolumn{1}{c|}{0.6781} &
  9.29 \\
MDDM\_G &
  0.8504 &
  0.4996 &
  0.7428 &
  0.5773 &
  0.7088 &
  0.7524 &
  0.6714 &
  0.5069 &
  0.6970 &
  0.6773 &
  0.6971 &
  \multicolumn{1}{c|}{0.6781} &
  9.21 \\
RDDM &
  \textBF{0.8696} &
  0.5140 &
  0.7417 &
  0.6560 &
  0.7654 &
  0.7669 &
  0.7847 &
  0.7227 &
  0.6970 &
  0.7272 &
  0.7117 &
  \multicolumn{1}{c|}{0.7286} &
  4.58 \\
SEQDRIFT2 &
  0.7942 &
  0.4996 &
  0.7421 &
  0.5773 &
  0.7088 &
  0.6304 &
  0.7763 &
  0.5069 &
  0.6970 &
  0.7266 &
  0.7327 &
  \multicolumn{1}{c|}{0.7778} &
  8.83 \\
ND &
  0.7953 &
  0.4996 &
  0.7632 &
  0.5773 &
  0.7088 &
  0.6685 &
  0.8168 &
  0.5069 &
  0.6970 &
  0.6773 &
  0.6971 &
  \multicolumn{1}{c|}{0.6781} &
  9.25 \\ \hline
Average Result &
  0.7744 &
  0.5006 &
  0.7475 &
  0.5825 &
  0.7144 &
  0.6332 &
  0.6845 &
  0.5392 &
  0.6995 &
  0.7078 &
  0.7150 &
  \multicolumn{1}{c|}{0.7109} &
  \textBF{Avg. Imp.} \\ \hline
LD3 Imp. (\%) &
  10.8 &
  45.9 &
  7.5 &
  37.0 &
  7.4 &
  26.7 &
  25.0 &
  36.1 &
  2.8 &
  14.4 &
  7.3 &
  \multicolumn{1}{c|}{16.8} &
  19.8 \\ \hline
\end{tabular}}
\end{subtable}

\end{minipage}
\end{center}
\end{table*}

\begin{table*}[]
\begin{center}
\begin{minipage}{\textwidth}
\ContinuedFloat
\caption{Continued.}
\begin{subtable}{1\textwidth}
\caption{Example-based F1 score and Micro-averaged F1 score results}\label{cst2}
\centering
\resizebox{\textwidth}{!}{%
\begin{tabular}{lccccccccccccr}
\hline
\textBF{Detector} &
  \textBF{20NG} &
  \textBF{Birds} &
  \textBF{Enron} &
  \textBF{\begin{tabular}[c]{@{}c@{}}Eukatyote\\ PseAAC\end{tabular}} &
  \textBF{Imdb} &
  \textBF{Ohsumed} &
  \textBF{\begin{tabular}[c]{@{}c@{}}Plant\\ PseAAC\end{tabular}} &
  \textBF{\begin{tabular}[c]{@{}c@{}}Tmc2007-\\ 500\end{tabular}} &
  \textBF{Yeast} &
  \textBF{\begin{tabular}[c]{@{}c@{}}Synthetic\\ Incremental\end{tabular}} &
  \textBF{\begin{tabular}[c]{@{}c@{}}Synthetic\\ Reoccurring\end{tabular}} &
  \multicolumn{1}{c|}{\textBF{\begin{tabular}[c]{@{}c@{}}Synthetic\\ Sudden\end{tabular}}} &
  \textBF{\begin{tabular}[c]{@{}r@{}}Average\\ Rank\end{tabular}} \\ \hline
\multicolumn{14}{c}{\textBF{Example-based F1 Score}} \\ \hline
LD3 &
  \textBF{0.3173} &
  0.1242 &
  \textBF{0.3496} &
  \textBF{0.4593} &
  \textBF{0.2240} &
  0.1031 &
  \textBF{0.5596} &
  \textBF{0.4048} &
  \textBF{0.5299} &
  0.1387 &
  \textBF{0.1719} &
  \multicolumn{1}{c|}{0.1325} &
  \textBF{2.75} \\
ADWIN &
  0.2999 &
  \textBF{0.1527} &
  0.3150 &
  0.1074 &
  0.2077 &
  0.0424 &
  0.2479 &
  0.3858 &
  0.5204 &
  0.1371 &
  0.1455 &
  \multicolumn{1}{c|}{0.1364} &
  7.46 \\
DDM &
  0.1207 &
  \textBF{0.1527} &
  0.3017 &
  0.1074 &
  0.1424 &
  0.0646 &
  0.2547 &
  0.3010 &
  0.5072 &
  0.1124 &
  0.1201 &
  \multicolumn{1}{c|}{0.1191} &
  12.08 \\
EDDM &
  0.1932 &
  \textBF{0.1527} &
  0.3155 &
  0.1074 &
  0.1606 &
  0.0557 &
  0.2737 &
  0.3458 &
  0.5159 &
  0.1236 &
  0.1340 &
  \multicolumn{1}{c|}{0.1265} &
  8.17 \\
FHDDM &
  0.2965 &
  \textBF{0.1527} &
  0.3050 &
  0.1074 &
  0.1409 &
  0.0424 &
  0.2408 &
  0.3024 &
  0.5204 &
  0.1210 &
  0.1556 &
  \multicolumn{1}{c|}{0.1211} &
  10.38 \\
FHDDMS &
  0.2915 &
  \textBF{0.1527} &
  0.3050 &
  0.1074 &
  0.1409 &
  0.0435 &
  0.2253 &
  0.3024 &
  0.5204 &
  0.1210 &
  0.1556 &
  \multicolumn{1}{c|}{0.1211} &
  10.25 \\
FHDDMS\_Add &
  0.3024 &
  \textBF{0.1527} &
  0.3077 &
  0.1074 &
  0.1409 &
  0.0439 &
  0.2283 &
  0.3024 &
  0.5204 &
  0.1210 &
  0.1556 &
  \multicolumn{1}{c|}{0.1211} &
  9.29 \\
HDDM\_A &
  0.3145 &
  \textBF{0.1527} &
  0.3176 &
  0.1074 &
  0.1409 &
  0.0530 &
  0.2549 &
  0.3024 &
  0.5204 &
  0.1264 &
  0.1556 &
  \multicolumn{1}{c|}{0.1386} &
  6.46 \\
HDDM\_W &
  0.3099 &
  \textBF{0.1527} &
  0.3176 &
  0.1074 &
  0.1409 &
  0.0526 &
  0.2104 &
  0.3024 &
  0.5204 &
  0.1210 &
  0.1556 &
  \multicolumn{1}{c|}{0.1211} &
  8.75 \\
KSWIN &
  0.2893 &
  \textBF{0.1527} &
  0.3030 &
  0.1074 &
  0.1409 &
  0.0974 &
  0.3273 &
  0.3024 &
  0.5204 &
  0.1210 &
  0.1556 &
  \multicolumn{1}{c|}{0.1211} &
  8.96 \\
MDDM\_A &
  0.2992 &
  \textBF{0.1527} &
  0.3066 &
  0.1074 &
  0.1409 &
  0.0426 &
  0.2422 &
  0.3024 &
  0.5204 &
  0.1210 &
  0.1556 &
  \multicolumn{1}{c|}{0.1211} &
  9.79 \\
MDDM\_E &
  0.3040 &
  \textBF{0.1527} &
  0.3065 &
  0.1074 &
  0.1409 &
  0.0428 &
  0.2375 &
  0.3024 &
  0.5204 &
  0.1210 &
  0.1556 &
  \multicolumn{1}{c|}{0.1211} &
  9.62 \\
MDDM\_G &
  0.3040 &
  \textBF{0.1527} &
  0.3065 &
  0.1074 &
  0.1409 &
  0.0428 &
  0.2375 &
  0.3024 &
  0.5204 &
  0.1210 &
  0.1556 &
  \multicolumn{1}{c|}{0.1211} &
  9.62 \\
RDDM &
  0.3021 &
  0.1461 &
  0.3029 &
  0.3905 &
  0.2239 &
  0.0429 &
  0.4141 &
  0.3986 &
  0.5204 &
  \textBF{0.1432} &
  0.1536 &
  \multicolumn{1}{c|}{\textBF{0.1436}} &
  6.54 \\
SEQDRIFT2 &
  0.2828 &
  \textBF{0.1527} &
  0.3241 &
  0.1074 &
  0.1409 &
  0.0429 &
  0.3099 &
  0.3024 &
  0.5204 &
  0.1299 &
  0.1410 &
  \multicolumn{1}{c|}{0.1339} &
  8.21 \\
ND &
  0.3099 &
  \textBF{0.1527} &
  0.3176 &
  0.1074 &
  0.1409 &
  \textBF{0.1044} &
  0.2430 &
  0.3024 &
  0.5204 &
  0.1210 &
  0.1556 &
  \multicolumn{1}{c|}{0.1211} &
  7.67 \\ \hline
Average Result &
  0.2813 &
  0.1523 &
  0.3102 &
  0.1263 &
  0.1523 &
  0.0543 &
  0.2632 &
  0.3108 &
  0.5192 &
  0.1241 &
  0.1500 &
  \multicolumn{1}{c|}{0.1259} &
  \textBF{Avg. Imp.} \\ \hline
LD3 Imp. (\%) &
  12.8 &
  -18.5 &
  12.7 &
  263.7 &
  47.1 &
  89.9 &
  112.6 &
  30.2 &
  2.1 &
  11.8 &
  14.6 &
  \multicolumn{1}{c|}{5.2} &
  48.7 \\ \hline
\multicolumn{14}{c}{\textBF{Micro-averaged F1 Score}} \\ \hline
LD3 &
  \textBF{0.2783} &
  \textBF{0.1804} &
  \textBF{0.2461} &
  \textBF{0.3258} &
  \textBF{0.2047} &
  \textBF{0.1297} &
  \textBF{0.4846} &
  \textBF{0.3374} &
  \textBF{0.5486} &
  \textBF{0.1462} &
  \textBF{0.1495} &
  \multicolumn{1}{c|}{\textBF{0.1461}} &
  \textBF{1.00} \\
ADWIN &
  0.2435 &
  0.1043 &
  0.2215 &
  0.0633 &
  0.1970 &
  0.0697 &
  0.2504 &
  0.3280 &
  0.5318 &
  0.1351 &
  0.1360 &
  \multicolumn{1}{c|}{0.1351} &
  8.42 \\
DDM &
  0.0997 &
  0.1043 &
  0.2230 &
  0.0633 &
  0.1480 &
  0.0702 &
  0.2024 &
  0.2652 &
  0.5247 &
  0.1393 &
  0.1411 &
  \multicolumn{1}{c|}{0.1393} &
  10.58 \\
EDDM &
  0.1227 &
  0.1043 &
  0.2387 &
  0.0633 &
  0.1641 &
  0.0711 &
  0.2257 &
  0.2749 &
  0.5409 &
  0.1460 &
  0.1417 &
  \multicolumn{1}{c|}{0.1385} &
  6.25 \\
FHDDM &
  0.2537 &
  0.1043 &
  0.2200 &
  0.0633 &
  0.1476 &
  0.0734 &
  0.2206 &
  0.2691 &
  0.5318 &
  0.1236 &
  0.1414 &
  \multicolumn{1}{c|}{0.1236} &
  9.92 \\
FHDDMS &
  0.2523 &
  0.1043 &
  0.2200 &
  0.0633 &
  0.1476 &
  0.0735 &
  0.2040 &
  0.2691 &
  0.5318 &
  0.1236 &
  0.1414 &
  \multicolumn{1}{c|}{0.1236} &
  10.29 \\
FHDDMS\_Add &
  0.2583 &
  0.1043 &
  0.2205 &
  0.0633 &
  0.1476 &
  0.0719 &
  0.2087 &
  0.2691 &
  0.5318 &
  0.1236 &
  0.1414 &
  \multicolumn{1}{c|}{0.1236} &
  9.96 \\
HDDM\_A &
  0.2616 &
  0.1043 &
  0.2411 &
  0.0633 &
  0.1476 &
  0.0708 &
  0.2199 &
  0.2691 &
  0.5318 &
  0.1306 &
  0.1414 &
  \multicolumn{1}{c|}{0.1346} &
  8.12 \\
HDDM\_W &
  0.2561 &
  0.1043 &
  0.2411 &
  0.0633 &
  0.1476 &
  0.0709 &
  0.1933 &
  0.2691 &
  0.5318 &
  0.1236 &
  0.1414 &
  \multicolumn{1}{c|}{0.1236} &
  9.75 \\
KSWIN &
  0.2514 &
  0.1043 &
  0.2214 &
  0.0633 &
  0.1476 &
  0.1115 &
  0.3041 &
  0.2691 &
  0.5318 &
  0.1236 &
  0.1414 &
  \multicolumn{1}{c|}{0.1236} &
  8.71 \\
MDDM\_A &
  0.2450 &
  0.1043 &
  0.2201 &
  0.0633 &
  0.1476 &
  0.0734 &
  0.2171 &
  0.2691 &
  0.5318 &
  0.1236 &
  0.1414 &
  \multicolumn{1}{c|}{0.1236} &
  10.38 \\
MDDM\_E &
  0.2485 &
  0.1043 &
  0.2209 &
  0.0633 &
  0.1476 &
  0.0734 &
  0.2111 &
  0.2691 &
  0.5318 &
  0.1236 &
  0.1414 &
  \multicolumn{1}{c|}{0.1236} &
  10.08 \\
MDDM\_G &
  0.2485 &
  0.1043 &
  0.2209 &
  0.0633 &
  0.1476 &
  0.0735 &
  0.2111 &
  0.2691 &
  0.5318 &
  0.1236 &
  0.1414 &
  \multicolumn{1}{c|}{0.1236} &
  9.88 \\
RDDM &
  0.2701 &
  0.1087 &
  0.2202 &
  0.2183 &
  0.2030 &
  0.0736 &
  0.3640 &
  0.3356 &
  0.5318 &
  0.1348 &
  0.1389 &
  \multicolumn{1}{c|}{0.1344} &
  5.33 \\
SeqDrift2 &
  0.2127 &
  0.1043 &
  0.2274 &
  0.0633 &
  0.1476 &
  0.0700 &
  0.2919 &
  0.2691 &
  0.5318 &
  0.1338 &
  0.1376 &
  \multicolumn{1}{c|}{0.1395} &
  9.25 \\
ND &
  0.2561 &
  0.1043 &
  0.2411 &
  0.0633 &
  0.1476 &
  0.1147 &
  0.2408 &
  0.2691 &
  0.5318 &
  0.1236 &
  0.1414 &
  \multicolumn{1}{c|}{0.1236} &
  8.08 \\ \hline
Average Result &
  0.2320 &
  0.1046 &
  0.2265 &
  0.0736 &
  0.1557 &
  0.0774 &
  0.2378 &
  0.2776 &
  0.5319 &
  0.1288 &
  0.1406 &
  \multicolumn{1}{c|}{0.1289} &
  \textBF{Avg. Imp.} \\ \hline
LD3 Imp. (\%) &
  20.0 &
  72.5 &
  8.7 &
  342.7 &
  31.5 &
  67.6 &
  103.4 &
  21.5 &
  3.1 &
  13.5 &
  6.3 &
  \multicolumn{1}{c|}{13.3} &
  58.7 \\ \hline
\end{tabular}%
}
\end{subtable}

\end{minipage}
\end{center}
\end{table*}


\par
The tests are conducted using prequential evaluation \citep{gama2009issues}. We apply the following four metrics to evaluate the effectiveness of the algorithms which are used in previous literature to measure multi-label classification effectiveness \citep{buyukccakir2018novel, nam2017maximizing}.
\begin{itemize}
    \item Example-based metrics: Example-based accuracy (Eq. \ref{eqn:accex}), Hamming score (Eq. \ref{eqn:hammscore}), and example-based F1 score (Eq. \ref{eqn:f1ex}). The formulas for these are given below in equations with $\hat{y}_i$ being the prediction, $Y_i$ as the ground truth, $n$ as the number of labels, and $N$ being the number of samples.
    \item Label-based metrics: Micro-averaged F1 score (Eq. \ref{eqn:f1mic}). $TP_i$, $FP_i$, and $FN_i$ are true positive, false positive, and false positive counts for the \emph{i-th} label \citep{zhang2013review}. 
\end{itemize}

\begin{equation}
Accuracy_{example} = \frac{1}{N} \sum^{N}_{i=1} \frac{\vert Y_i \bigcap \hat{y}_i\vert }{\vert Y_i \bigcup \hat{y}_i\vert}
\label{eqn:accex}
\end{equation}

\begin{equation}
Hamming Score = 1 - \frac{1}{N} \sum^{N}_{i=1} \frac{1}{n} \vert \hat{y}_i \Delta Y_i\vert
\label{eqn:hammscore}
\end{equation}

\begin{equation} 
F1_{example}\footnote{$Precision_{example} = \frac{1}{N} \sum^{N}_{i=1} \frac{\vert Y_i \bigcap \hat{y}_i \vert}{\vert \hat{y}_i\vert}$ \\ $Recall_{example} = \frac{1}{N} \sum^{N}_{i=1} \frac{\vert Y_i \bigcap \hat{y}_i \vert}{\vert Y_i\vert}$} = \frac{2 \cdot Precision_{example} \cdot Recall_{example}}{Precision_{example} + Recall_{example}}
\label{eqn:f1ex}
\end{equation} 


\setlength{\footnotesep}{0.5\baselineskip}

\begin{equation}
F1_{micro}\footnote{$F1(TP, FP, FN)=\frac{2 \cdot TP}{2 \cdot TP + FN + FP}$} = F1 \left( \sum^{n}_{i=1} TP_i, \sum^{n}_{i=1} FP_i, \sum^{n}_{i=1} FN_i\right)
\label{eqn:f1mic}
\end{equation}

LD3 is compared with the detection algorithms displayed in Table \ref{tab:detector-table} and their variations. Although these detection algorithms are not specifically designed for multi-label use-case, they are eligible baseline algorithms because they are error rate-based detectors. For this type of algorithms, the input passed into the detector is whether or not the prediction is correct, which means that their input is ``1" for correct and ``0" for false predictions (For ``1", all predicted labels must match exactly to the true labels). For this reason, they can also be used in the multi-label concept detection problem domain since this information is also generated in multi-label data streams. \cite{wang2020concept} also test their multi-label concept drift detection algorithm against DDM.



\section{Experimental Results and Discussion}\label{sec6}
In the following subsections, we discuss and analyze our results based on effectiveness measures, provide simple guidelines for hyperparameter optimization, discuss the effects of different data fusion algorithms, and the influence of various drift types and speeds.

\subsection{Effectiveness Analysis}
We assess our effectiveness results by comparing our experimental results with the baseline detectors. Furthermore, we visually analyze the effectiveness results in the time scale. Lastly, we discuss the advantages of unsupervised detection with pointers on efficiency.

\subsubsection{Analysis with Different Effectiveness Measures}
The results of our experiments are presented in Table \ref{tab:results}. Each row represents the results of a detector and columns represent the datasets. The table is divided into four sections for each effectiveness measure. The baseline classifier without a detector is labeled as ``ND", i.e, no detector. While calculating the average rankings, the ties are handled by averaging the ranks that would be assigned to all the tied values and assigning that value to each tied element, e.g., for the given list $[1,2,3,3,4]$, the average ranking is $[1,2,3.5,3.5,5]$.

\par

The results show that LD3 achieves the best results overall on all metrics. For all metrics, LD3 achieves at least 19.8\% improvement over the baseline averages. However, average rank-wise, LD3 displays comparatively worse predictive performance on the example-based F1 score than other metrics. We found that the reason behind this is mostly due to the classifier resetting procedure. Example-based F1 score punishes miss-classifications more harshly than example-based accuracy since miss-classified samples are multiplied in its numerator and they are divided by additive components. A recently reset classifier is prone to incorrect predictions and it is usually the case that none of the predicted labels match with the true labels for the first samples after resetting which produces the outcome of worse example-based F1 scores. 

\begin{figure*}[!ht]
\centering
\begin{framed}
\begin{subfigure}{\textwidth}
  \centering
  \includegraphics[width=\textwidth]{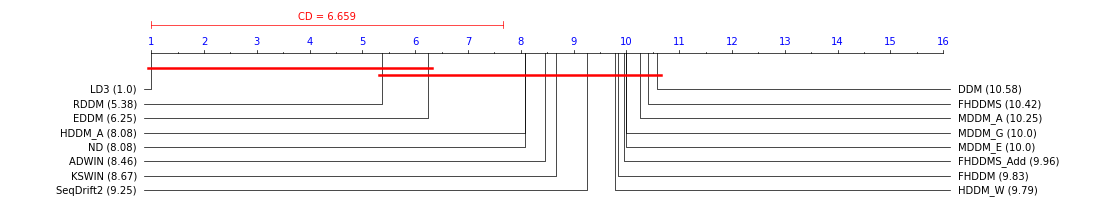}
  \caption{Example-based accuracy}
  \label{fig:sub1}
\end{subfigure}
\begin{subfigure}{\textwidth}
  \centering
  \includegraphics[width=\textwidth]{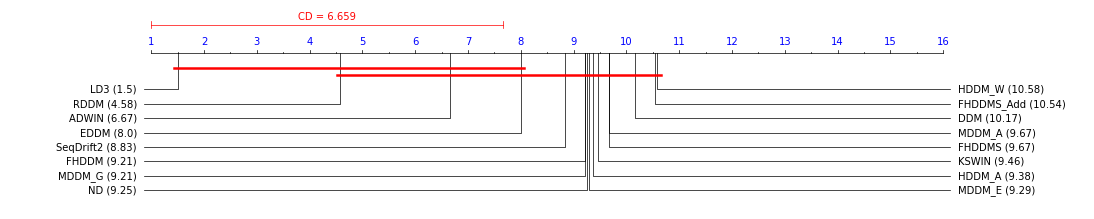}
  \caption{Hamming score}
  \label{fig:sub2}
\end{subfigure}  
\begin{subfigure}{\textwidth}
  \centering
  \includegraphics[width=\textwidth]{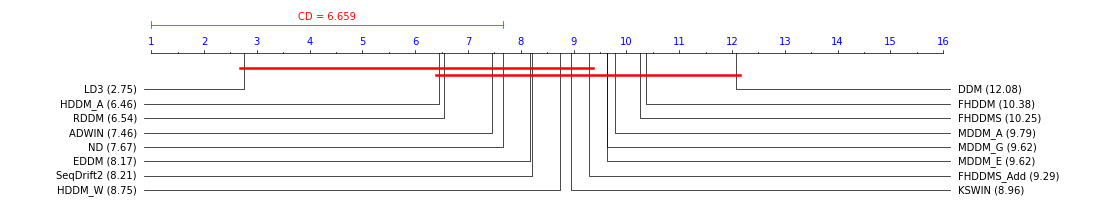}
  \caption{Example-based F1 score}
  \label{fig:sub1}
\end{subfigure}
\begin{subfigure}{\textwidth}
  \centering
  \includegraphics[width=\textwidth]{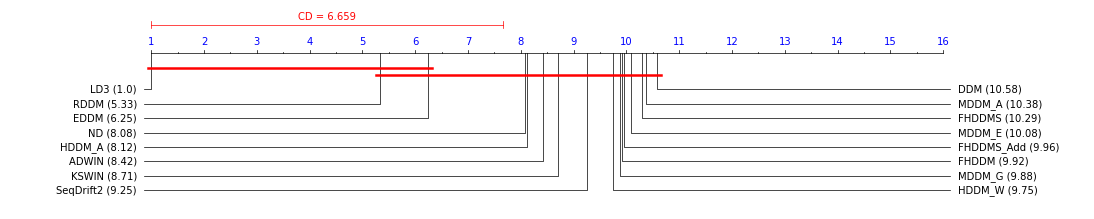}
  \caption{Micro-averaged F1 score}
  \label{fig:sub1}
\end{subfigure}
\end{framed}
\caption{Nemenyi critical distance diagrams for all metrics. The number given within parentheses after the method name indicates rank position of the method.}
\label{fig:nemenyi}
\end{figure*}

\subsubsection{Statistical Significance Evaluation of Effectiveness Results}
To further investigate the effectiveness of LD3, we performed the \emph{``Friedman test with Nemenyi post-hoc analysis"} in which we evaluate the statistical significance of our results, which is presented in Figure \ref{fig:nemenyi}. We applied the two-tailed Nemenyi test to find our critical distance for Nemenyi Significance. Our critical distance is $CD = 6.659$, which is calculated using Equation \ref{eqn:nemenyi}, where $q_{\alpha,k}$ is the critical value acquired from the Critical Values Table from \cite{demvsar2006statistical} with $\alpha=0.05$ and $k$ is the number of algorithms (16) and $K$ is the number of datasets (12).

\begin{equation}
CD=q_{\alpha, k}\sqrt{\frac{k(k+1)}{6K}}
\label{eqn:nemenyi}
\end{equation}

\par
The Nemenyi Significance tests show that LD3 achieves the overall best rankings with a statistically significant difference from most of the detectors. Among the different effectiveness measures, only RDDM and EDDM consistently display statistically indistinguishable predictive performance.

\begin{figure*}[bp]
\centering
\begin{subfigure}{0.8\columnwidth}
  \centering
  \includegraphics[width=\columnwidth]{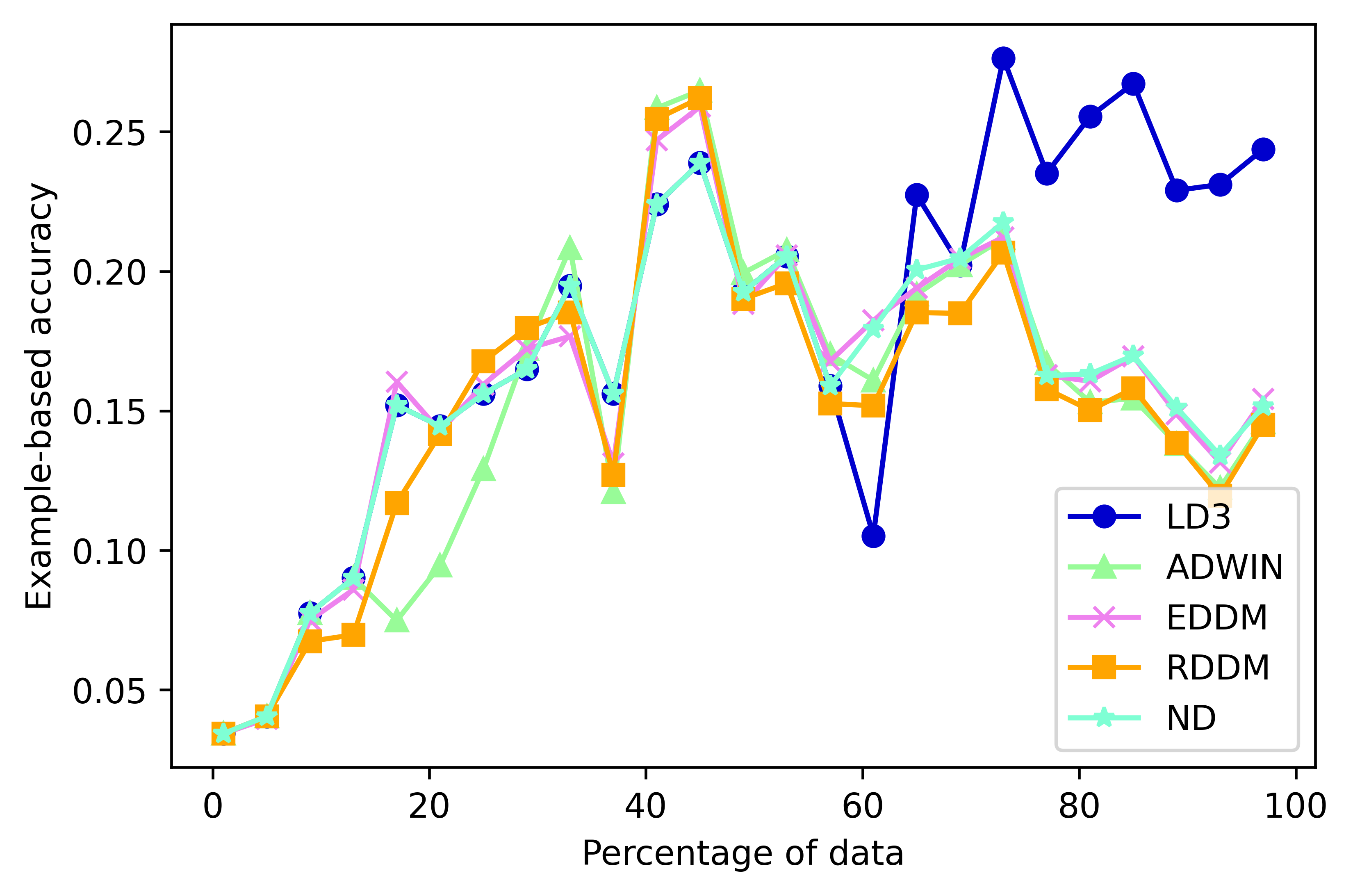}
  \caption{Enron}
  \label{fig:sub11}
\end{subfigure}%
\begin{subfigure}{0.8\columnwidth}
  \centering
  \includegraphics[width=\columnwidth]{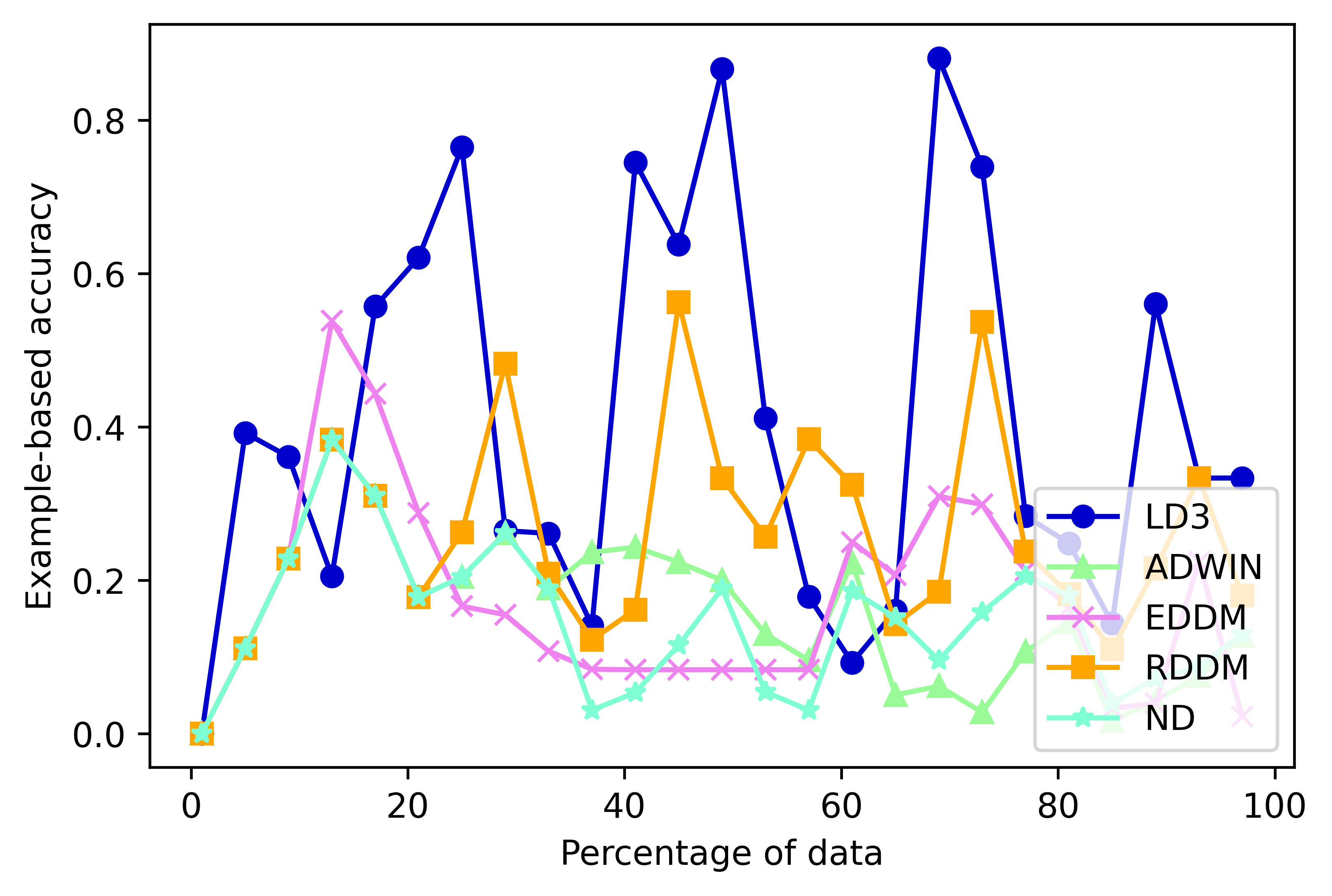}
  \caption{PlantPseAAC}
  \label{fig:sub12}
\end{subfigure} \hspace{50mm}
\begin{subfigure}{0.8\columnwidth}
  \centering
  \includegraphics[width=\columnwidth]{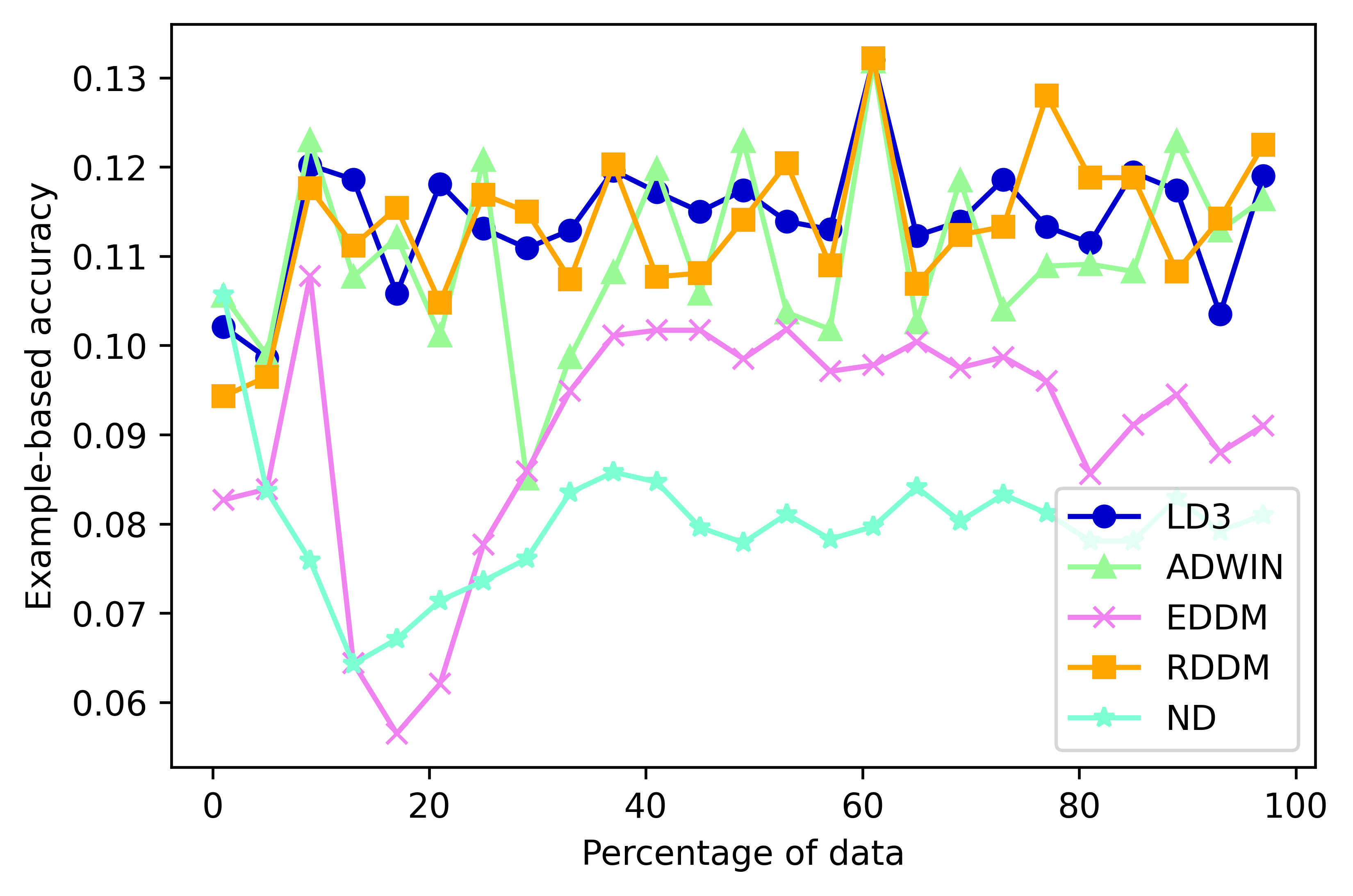}
  \caption{Imdb}
  \label{fig:sub13}
\end{subfigure}%
\begin{subfigure}{0.8\columnwidth}
  \centering
  \includegraphics[width=\columnwidth]{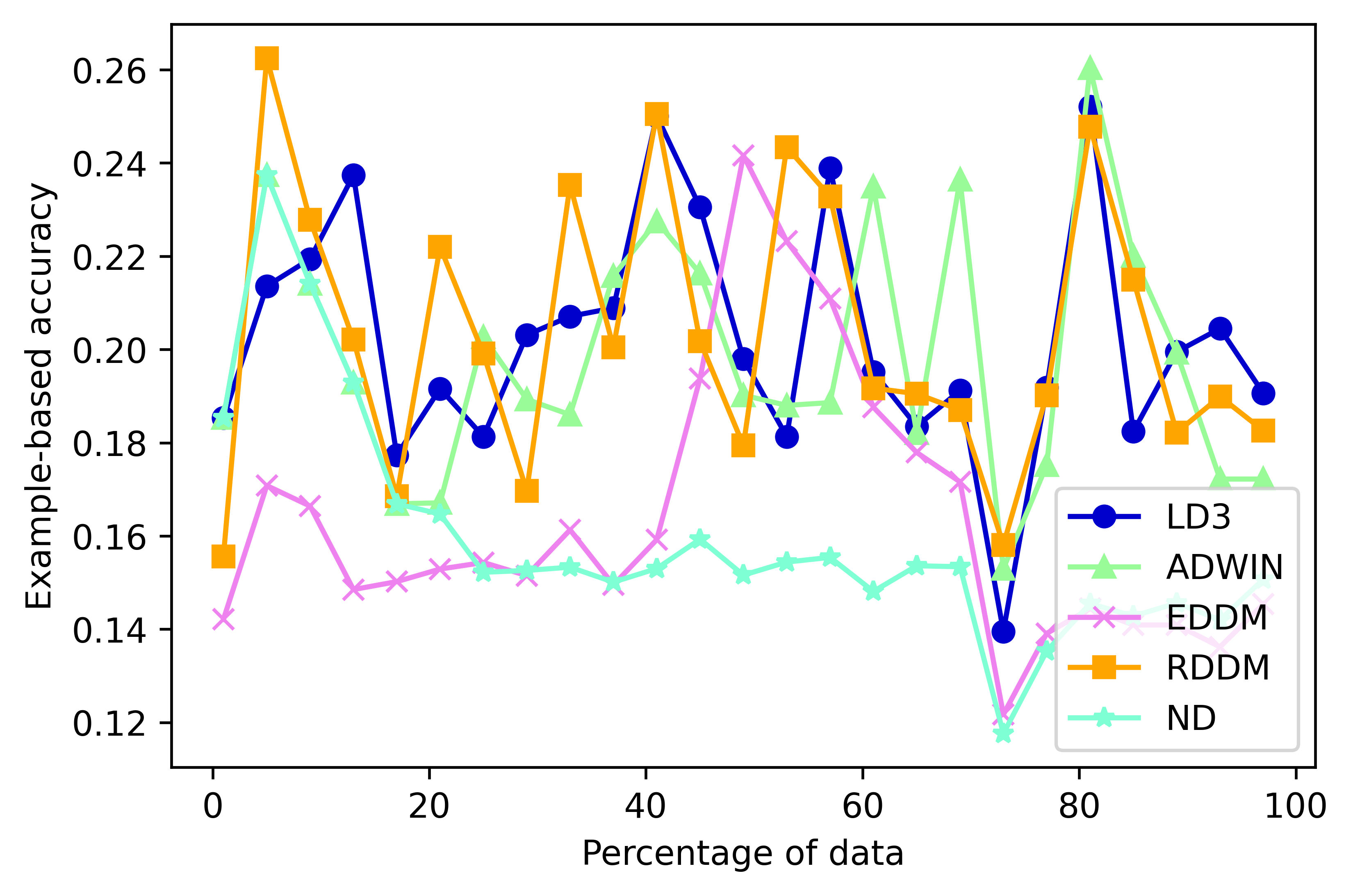}
  \caption{Tmc2007-500}
  \label{fig:sub14}
\end{subfigure}

\caption{Example-based accuracy results of the top detectors and ND on four datasets. Y-axis of the plots is given with different scales.}
\label{fig:acc-comp}
\end{figure*}

It should be observed that contrary to prior expectation, ND does not show the worst results and it achieves better results compared to more than half of the tested detectors. Our experiments show that apart from ADWIN, EDDM, HDDM\_A, and RDDM, the baseline detectors either do not detect drift or perform worse than ND in general. We identify the reason behind this as frequent false positive drift detections. An incorrectly detected drift leads the classifier to discard the progress it made which severely reduces the effectiveness results. In addition, after a reset, the error rate of the classifier is increased which causes an error rate-based detector to miss-interpret the miss-classification statistics and may result in oscillating drift detections. The positive feedback loop generated from both of these factors causes the detectors to produce worse results than ND.

\par

However, our experiments confirm that ADWIN, EDDM, HDDM\_A, and RDDM are suitable for drift detection on multi-label data streams. For the following discussions, we use the top four detectors in terms of average rank, namely, LD3, ADWIN, EDDM, and RDDM.

\subsubsection{Time Change Analysis of Effectiveness}
One aspect we would like to highlight is LD3's ability to detect drifts even when tested with a dataset that has low label cardinality. Our initial assumption was that LD3 would perform worse on datasets with lower label cardinality, since there would be less co-occurrence. However, in datasets with low label cardinality such as Birds, 20NG, and PlantPseAAC; LD3 still outperforms the baselines in general. We found that even if the label cardinality is low, as long as the data stream is a multi-label stream, some labels still have more influence over other labels and LD3 continues to be effective. One problem that arises from this is that how the algorithm should respond if the nature of the stream changes from multi-label to multi-class stream. Our suggestion in such a case would be to switch to a different detector suited to multi-class streams since LD3 would lose its ability to detect drifts. However, we were unable to find any detectors that could detect a change in the data stream's nature and recommend further research for such cases.
\par

Figure \ref{fig:acc-comp} plots the time scale example-based accuracy results of the top four detectors and ND. These datasets are chosen as most of the detectors co-detect drifts and their plots are more clear, allowing better inspection. The datasets are divided into 25 subsets and the average accuracy within those subsets is plotted.  
\par
Most of the sudden decreases in accuracy are caused by the classifier resets which means that at those points, the algorithms deduce that drift is present. Noticeably on the PlantPseAAC dataset, the sudden decreases in accuracy occur earlier for LD3, after which the classifier produces higher accuracy values than the baselines. LD3's early detections allow the classifier to learn larger amounts of samples and adapt to the new concept faster. This is also observed in the second half of the Enron dataset where the other detectors stagnate in accuracy while LD3 resets to learn the new drift.

\begin{figure*}[tp]
\centering
\begin{subfigure}{.8\columnwidth}
  \centering
  \includegraphics[width=.8\columnwidth]{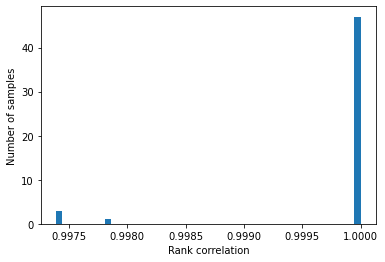}
  \caption{EukaryotePseAAC with $t = 2$}
  \label{fig:sub1}
\end{subfigure}%
\begin{subfigure}{.8\columnwidth}
  \centering
  \includegraphics[width=.8\columnwidth]{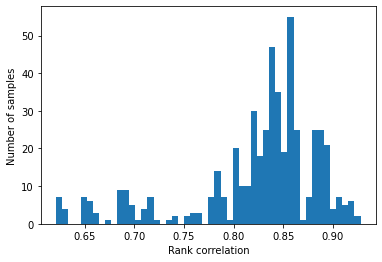}
  \caption{Synthetic Sudden with $t = 3$}
  \label{fig:sub2}
\end{subfigure} \hspace{50mm}
\begin{subfigure}{.8\columnwidth}
  \centering
  \includegraphics[width=.8\columnwidth]{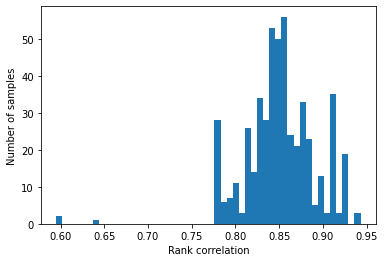}
  \caption{Tmc2007-500 with $t = 4$}
  \label{fig:sub1}
\end{subfigure}%
\begin{subfigure}{.8\columnwidth}
  \centering
  \includegraphics[width=.8\columnwidth]{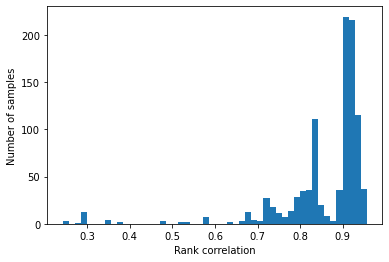}
  \caption{20NG with $t = 5$}
  \label{fig:sub1}
\end{subfigure}
\caption{Distribution of the rank correlations within $W_{corr}$ of four datasets after a drift is detected with different $t$ values ($\sigma$ multipliers). X-axis of the plots is given with different scales.}
\label{fig:hyperparameter}
\end{figure*}

Furthermore, the plots illustrate that LD3 is more resistant to the noisy data originating from a recently reset classifier. Notably with ADWIN and RDDM on Tmc2007-500 and Imdb datasets, we observe that there are numerous consecutive drops in accuracy, which is the result of a new detection following a recent reset. While the classifier builds a probabilistic model for the new distribution, some noise is usually expected. Yet for error rate-based algorithms like ADWIN and RDDM, this noise causes false detections that result in oscillating resets as previously discussed. LD3 combats this problem by waiting until sufficient statistics are obtained while the windows get filled, which allows faster recovery. For instance, in the Tmc2007-500 dataset, RDDM generally reaches higher accuracy values. However, RDDM resets the classifier frequently as opposed to LD3's more stable execution, which results in marginally worse overall accuracy (LD3: 0.2029, RDDM: 0.2016) for RDDM, despite having multiple higher local maximum values. A similar case is also displayed with the Imdb dataset.
\par
We define robustness, in the context of concept drift detection, as the ability to aid classifiers to provide the best predictive performance within a data stream with changing concepts. Our investigation on Figure \ref{fig:acc-comp} and results on Table \ref{tab:results} demonstrate that LD3 is a robust detection algorithm that assists the classifier to achieve the best effectiveness results overall, within varying streaming environments with drift.

\subsubsection{Efficiency Concerns and Advantages of Unsupervised Detection}

In our study, we chose not to include an efficiency analysis, because, in our experiments, we observe that a drift detector's influence on the execution time is much lower than the paired classifier. Furthermore, supervised detection algorithms assume that the true labels are readily available \citep{sethi2017reliable, vzliobaite2010change}; however, that is not always the case in real life. The acquisition of labeled data is usually difficult and it is much easier to collect unlabeled data \citep{subhashini2021mining}. As LD3 is an unsupervised detector, the lack of labeled data does not prevent its continuous concept drift detection whereas supervised algorithms may halt while waiting for labels.

\subsection{Hyperparameter Analysis}
In our analysis of the hyperparameters, we propose a method to set the standard deviation multiplier $t$. Furthermore, we conduct experiments to determine the effect of window size $w$ and threshold for the number of anomaly $L$. 
\subsubsection{Setting the Standard Deviation Multiplier $t$}
We developed a simple way to tune the $t$ parameter of LD3. Figure \ref{fig:hyperparameter} plots the distribution of past correlations within $W_{corr}$, when a drift is detected for four datasets as histogram plots.

Since LD3 is an unsupervised detector, such plots could easily be obtained as a part of the test runs of a given data stream. In the figure, we see that with each subplot the range of correlation values expand; however the mean correlation does not decrease at the same rate, where they stay in $[0.80, 1]$. If the standard deviation of the correlations are very small, e.g, EukaryotePseAAC dataset in Figure \ref{fig:hyperparameter}.a, a low $t$ is required such as $t = 2$. With this $t$ value, if the current correlation value is smaller than $\mu - 2\sigma$, it lies outside of the 95\% confidence interval and it should be considered as an anomaly \citep{pukelsheim1994three}. For an accurate tuning of LD3, starting from $t = 2$, the confidence interval should be expanded with an increasing $t$, with guidance from the plots generated, similar to the examples in Figure \ref{fig:hyperparameter}. We used this method to determine the $t$ values for our experiments on Table \ref{tab:results}. Among the parameter values we used, $t = 4$ most frequently achieves the best effectiveness result. Therefore, we use this value as a default parameter.


\subsubsection{Setting the Parameters Window Size $w$ and Threshold for Number of Anomalies $L$}

To investigate the effect of $L$ and $w$ parameters, we conducted experiments on the real-world datasets using $w = [50,250,500]$, $L = [0,1,2]$ values. Table \ref{tab:param-setting} presents the results of our experiments. We found that $L=0$ is a good default value since most of the best results are achieved with this value. This indicates that our concerns about outlier resiliency are not as significant as initially assumed. Though this is most likely not the case for every data stream as streams with extreme noise may be susceptible to outliers. We urge researchers who would utilize LD3 to better tune the $L$ parameter if there are available resources such as labeled data and computational capacity.

\begin{table*}[tp]
\centering
\caption{Example-based accuracy of LD3 with varying parameter settings for each real-world dataset The best scores are highlighted in bold. Datasets are listed in ascending order of the number of data items they contain.}
\label{tab:param-setting}
\resizebox{\textwidth}{!}{%
\begin{tabular}{l|r|ccccccccc}
\hline
\textBF{Dataset} &
  \multicolumn{1}{c|}{\boldmath$N$} &
  \textBF{\begin{tabular}[c]{@{}c@{}}\boldmath$w=50$\\ \boldmath$L=0$\end{tabular}} &
  \textBF{\begin{tabular}[c]{@{}c@{}}\boldmath$w=50$\\ \boldmath$L=1$\end{tabular}} &
  \textBF{\begin{tabular}[c]{@{}c@{}}\boldmath$w=50$\\ \boldmath$L=2$\end{tabular}} &
  \textBF{\begin{tabular}[c]{@{}c@{}}\boldmath$w=250$\\ \boldmath$L=0$\end{tabular}} &
  \textBF{\begin{tabular}[c]{@{}c@{}}\boldmath$w=250$\\ \boldmath$L=1$\end{tabular}} &
  \textBF{\begin{tabular}[c]{@{}c@{}}\boldmath$w=250$\\ \boldmath$L=2$\end{tabular}} &
  \textBF{\begin{tabular}[c]{@{}c@{}}\boldmath$w=500$\\ \boldmath$L=0$\end{tabular}} &
  \textBF{\begin{tabular}[c]{@{}c@{}}\boldmath$w=500$\\ \boldmath$L=1$\end{tabular}} &
  \textBF{\begin{tabular}[c]{@{}c@{}}\boldmath$w=500$\\ \boldmath$L=2$\end{tabular}} \\ \hline
Birds &
  645 &
  \textBF{0.0618} &
  0.0550 &
  0.0550 &
  0.0550 &
  0.0550 &
  0.0550 &
  0.0550 &
  0.0550 &
  0.0550 \\
PlantPseAAC &
  978 &
  \textBF{0.1383} &
  0.1369 &
  0.1369 &
  0.1369 &
  0.1369 &
  0.1369 &
  0.1369 &
  0.1369 &
  0.1369 \\
Enron &
  1,702 &
  0.1055 &
  0.1089 &
  0.1371 &
  \textBF{0.1403} &
  0.1301 &
  0.1371 &
  0.1371 &
  0.1371 &
  0.1371 \\
Yeast &
  2,417 &
  \textBF{0.3780} &
  0.3644 &
  0.3622 &
  0.3724 &
  0.3720 &
  0.3718 &
  0.3613 &
  0.3613 &
  0.3613 \\
EukaryotePseAAC &
  7,766 &
  \textBF{0.1663} &
  0.0727 &
  0.0327 &
  0.0627 &
  0.0647 &
  0.0658 &
  0.0587 &
  0.0578 &
  0.0593 \\
Ohsumed &
  13,930 &
  0.0367 &
  0.0370 &
  0.0608 &
  0.0444 &
  0.0443 &
  0.0443 &
  \textBF{0.0630} &
  \textBF{0.0630} &
  0.0527 \\
20NG &
  19,300 &
  0.0667 &
  0.0733 &
  0.1469 &
  0.1209 &
  0.1219 &
  0.1247 &
  \textBF{0.1516} &
  0.1384 &
  0.1497 \\
Tmc2007-500 &
  28,600 &
  0.1485 &
  0.1918 &
  0.1555 &
  0.1931 &
  0.1934 &
  0.1980 &
  \textBF{0.2029} &
  0.2009 &
  0.2007 \\
Imdb &
  120,900 &
  0.0717 &
  0.0810 &
  0.0797 &
  0.0950 &
  0.0954 &
  0.1012 &
  0.1129 &
  0.1124 &
  \textBF{0.1140} \\ \hline
\end{tabular}%
}
\end{table*}

Unlike the $L$ parameter, our results show that setting the $w$ parameter to a default value is not as straightforward. In Table \ref{tab:param-setting} we observe that the datasets with low number of samples performs best with $w=50$ and larger datasets with $w=500$. This is related to the problem of sampling rate for data streams. Depending on the problem domain, the source of a stream can be sampled daily for stable streams such as music billboard charts and it can be sampled several times a minute for volatile data streams such as the price of a cryptocurrency. For this purpose it is possible to use a hyperparameter self-tuning scheme \citep{veloso2021hyperparameter}, to better adjust the classifier based on the current distribution. In our experiments, we observe that the value of the $w$ parameter is highly dependant on the sampling rate of a stream and it should be set accordingly. For frequently sampled data streams we recommend $w=500$ as a default value and $w=50$ for infrequently sampled streams.

\subsection{Analysis of Different Fusion Methods}

In order to study the effects of different data fusion algorithms and to choose a suitable, default data fusion algorithm for LD3, we conducted experiments on the nine real-world datasets we used on effectiveness tests. We chose to exclude the synthetic data streams for these experiments as we aim to evaluate the general effectiveness of the tested data fusion algorithms in real-world datasets. In these experiments, in addition to reciprocal ranking, we used popular data fusion algorithms Borda Fuse, Condorcet's Fuse \citep{nuray2006automatic} and Markov Chains Type 4 (MC4) \citep{dwork2001rank}. The results are shown in Table \ref{tab:data-fusion-type}. The tests are done by only changing the data fusion section of LD3.

\begin{table}[!h]
\centering
\caption{Example-based accuracy results of LD3 using four different data fusion algorithms. Best results are highlighted in bold.}
\label{tab:data-fusion-type}
\resizebox{\columnwidth}{!}{\begin{tabular}{lcccc}
\hline
\textBF{Dataset} & \textBF{Reciprocal} & \textBF{Borda}  & \textBF{Condorcet} & \textBF{MC4} \\ \hline
20NG             & \textBF{0.1616}     & 0.1480          & 0.1497             & 0.1517       \\
Birds            & \textBF{0.0992}     & 0.0913          & 0.0984             & 0.0903       \\
Enron            & \textBF{0.1403}     & 0.1343          & 0.1316             & 0.1368       \\
EukaryotePseAAC  & 0.1946              & \textBF{0.2096} & 0.0727             & 0.0685       \\
Imdb             & 0.1140              & \textBF{0.1154} & 0.1110             & 0.1134       \\
Ohsumed          & \textBF{0.0693}     & 0.0568          & 0.0595             & 0.0594       \\
PlantPseAAC      & 0.3198              & \textBF{0.3393} & 0.2397             & 0.2099       \\
Tmc2007-500      & \textBF{0.2029}     & 0.1954          & 0.1999             & 0.1997       \\
Yeast            & \textBF{0.3765}     & 0.3643          & 0.3577             & 0.3586       \\
\hline
\end{tabular}}
\end{table}

The results show that reciprocal rank fusion provides better results (in six of the nine cases). This is within our expectations as past studies such as \cite{cormack2009reciprocal} and \cite{pedronette2015unsupervised} have shown similar observations in different problem domains.

\subsection{Influence of Drift Types and Speed}
Regarding drift types, since LD3 counts the number of co-occurrences to detect concept drift, incremental, sudden, and reoccurring drifts have the same structure as all of these types of drift result in changing rankings. As shown in Table \ref{tab:results}, LD3 generally performs the same between different types of drifts and outperforms other detectors. Also, from the algorithm perspective, both incremental and gradual drifts show the same structure within the co-occurrence matrix. Therefore, we chose not to add an additional stream for gradual drift as the results were very similar.

\begin{table}[!h]
\centering
\caption{Example-based accuracy results of the top detectors with different drift types. Best results are highlighted in bold.}
\label{tab:drift-speed}
\resizebox{\columnwidth}{!}{\begin{tabular}{lccccc}
\hline
\textBF{Data Stream} & \textBF{LD3}    & \textBF{ADWIN} & \textBF{EDDM} & \textBF{RDDM} & \textBF{ND} \\ \hline
Sudden\_1        & \textBF{0.0788} & 0.0724         & 0.0744           & 0.0720        & 0.0659      \\
Sudden\_25       & \textBF{0.0767} & 0.0724         & 0.0741           & 0.0724        & 0.0659      \\
Sudden\_50       & \textBF{0.0776} & 0.0725         & 0.0737           & 0.0724        & 0.0659      \\
Incremental\_250     & \textBF{0.0769} & 0.0725         & 0.0733           & 0.0731        & 0.0659      \\
Incremental\_500     & \textBF{0.0788} & 0.0724         & 0.0787           & 0.0722        & 0.0659      \\
Incremental\_1000    & \textBF{0.0776} & 0.0717         & 0.0766           & 0.0713        & 0.0659      \\ \hline
\end{tabular}}
\end{table}

In addition to drift types, the speed at which the drift occurs also influences a detector's ability to find drifts. In Table \ref{tab:drift-speed}, the accuracy results of the top four detectors are presented. The streams used are synthetic data streams with sudden and incremental drifts and the number after the underscore is the number of samples over which the drift occurs. We generate them using the Scikit-Multiflow framework \citep{montiel2018scikit} by the same method described in Section 5.1 where they have 20,000 samples, 50 classes, and 200 features with drifts occurring at sample positions 4,000 and 10,000. Our findings indicate that regardless of the drift speed and type, LD3 still outperforms the baseline detectors, which demonstrate LD3's robustness under different speed conditions.

\section{Conclusion}\label{sec7}
In this paper, we present LD3, a novel unsupervised concept drift detection algorithm that exploits dynamic temporal dependencies between class labels in a multi-label classification environment. The algorithm is based on a new concept, label dependency ranking, which we introduce. We perform an extensive evaluation of LD3 against 14 prevalent drift detection algorithms using a Classifier Chain of Gaussian Naive Bayes classifiers. The experimental results show that LD3 provides the highest classifier accuracy: Without using true class labels, it statistically significantly outperforms several of the other drift detection algorithms that require such labels. 
\par
Our algorithm is able to accurately assist multi-label classifiers in the presence of a concept drift, which results in more robust classification models that are more adaptive to changing trends. In many streaming environments, true class labels required by supervised concept drift detection methods are likely to be unavailable due to several reasons. As LD3 is an unsupervised algorithm, this advantage illustrates the practical value of LD3.
\par
In future work, we plan to examine the uses of LD3 in unsupervised classification and in environments that include concept evolution, i.e. the emergence of entirely new labels.

\section*{Acknowledgements}
This study is partially supported by Turkcell İletişim Hizmetleri A.Ş. within the framework of 5G and Beyond Joint Graduate Support Programme coordinated by Information and Communication Technologies Authority.

\balance
\bibliography{sn-article}

\begin{thebibliography}{58}
\providecommand{\natexlab}[1]{#1}
\providecommand{\url}[1]{{#1}}
\providecommand{\urlprefix}{URL }
\providecommand{\doi}[1]{\url{https://doi.org/#1}}
\providecommand{\eprint}[2][]{\url{#2}}
 \bibcommenthead

\bibitem[{Baena-Garc{\i}a et~al(2006)Baena-Garc{\i}a, del Campo-{\'A}vila,
  Fidalgo, Bifet, Gavalda, and Morales-Bueno}]{baena2006early}
Baena-Garc{\i}a M, del Campo-{\'A}vila J, Fidalgo R, et~al (2006) Early drift
  detection method. In: Fourth International Workshop on Knowledge Discovery
  from Data Streams, pp 77--86

\bibitem[{Bahri et~al(2021)Bahri, Bifet, Gama, Gomes, and
  Maniu}]{bahri2021data}
Bahri M, Bifet A, Gama J, et~al (2021) Data stream analysis: Foundations, major
  tasks and tools. Wiley Interdisciplinary Reviews: Data Mining and Knowledge
  Discovery 11(3):e1405

\bibitem[{Barros et~al(2017)Barros, Cabral, Gon{\c{c}}alves~Jr, and
  Santos}]{barros2017rddm}
Barros RS, Cabral DR, Gon{\c{c}}alves~Jr PM, et~al (2017) Rddm: Reactive drift
  detection method. Expert Systems with Applications 90:344--355

\bibitem[{Bifet and Gavalda(2007)}]{bifet2007learning}
Bifet A, Gavalda R (2007) Learning from time-changing data with adaptive
  windowing. In: Proceedings of the 2007 SIAM International Conference on Data
  Mining, SIAM, pp 443--448

\bibitem[{Bonab and Can(2018)}]{bonab2018goowe}
Bonab HR, Can F (2018) {GOOWE}: Geometrically optimum and online-weighted
  ensemble classifier for evolving data streams. ACM Transactions on Knowledge
  Discovery from Data (TKDD) 12(2):1--33

\bibitem[{B{\"u}y{\"u}k{\c{c}}akir et~al(2018)B{\"u}y{\"u}k{\c{c}}akir, Bonab,
  and Can}]{buyukccakir2018novel}
B{\"u}y{\"u}k{\c{c}}akir A, Bonab H, Can F (2018) A novel online stacked
  ensemble for multi-label stream classification. In: Proceedings of the 27th
  ACM International Conference on Information and Knowledge Management, pp
  1063--1072

\bibitem[{Cormack et~al(2009)Cormack, Clarke, and
  Buettcher}]{cormack2009reciprocal}
Cormack GV, Clarke CL, Buettcher S (2009) Reciprocal rank fusion outperforms
  condorcet and individual rank learning methods. In: Proceedings of the 32nd
  International ACM SIGIR Conference on Research and Development in Information
  Retrieval, pp 758--759

\bibitem[{Dem{\v{s}}ar(2006)}]{demvsar2006statistical}
Dem{\v{s}}ar J (2006) Statistical comparisons of classifiers over multiple data
  sets. Journal of Machine Learning Research 7(Jan):1--30

\bibitem[{Dwork et~al(2001)Dwork, Kumar, Naor, and Sivakumar}]{dwork2001rank}
Dwork C, Kumar R, Naor M, et~al (2001) Rank aggregation methods for the web.
  In: Proceedings of the 10th International Conference on World Wide Web, pp
  613--622

\bibitem[{Fr{\'\i}as-Blanco et~al(2014)Fr{\'\i}as-Blanco, del Campo-{\'A}vila,
  Ramos-Jimenez, Morales-Bueno, Ortiz-D{\'\i}az, and
  Caballero-Mota}]{frias2014online}
Fr{\'\i}as-Blanco I, del Campo-{\'A}vila J, Ramos-Jimenez G, et~al (2014)
  Online and non-parametric drift detection methods based on hoeffding’s
  bounds. IEEE Transactions on Knowledge and Data Engineering 27(3):810--823

\bibitem[{Gama et~al(2004)Gama, Medas, Castillo, and
  Rodrigues}]{gama2004learning}
Gama J, Medas P, Castillo G, et~al (2004) Learning with drift detection. In:
  Brazilian Symposium on Artificial Intelligence, Springer, pp 286--295

\bibitem[{Gama et~al(2009)Gama, Sebasti{\~a}o, and Rodrigues}]{gama2009issues}
Gama J, Sebasti{\~a}o R, Rodrigues PP (2009) Issues in evaluation of stream
  learning algorithms. In: Proceedings of the 15th ACM SIGKDD International
  Conference on Knowledge Discovery and Data Mining, pp 329--338

\bibitem[{Gama et~al(2014)Gama, {\v{Z}}liobait{\.e}, Bifet, Pechenizkiy, and
  Bouchachia}]{gama2014survey}
Gama J, {\v{Z}}liobait{\.e} I, Bifet A, et~al (2014) A survey on concept drift
  adaptation. ACM Computing Surveys (CSUR) 46(4):1--37

\bibitem[{Gemaque et~al(2020)Gemaque, Costa, Giusti, and
  Dos~Santos}]{gemaque2020overview}
Gemaque RN, Costa AFJ, Giusti R, et~al (2020) An overview of unsupervised drift
  detection methods. Wiley Interdisciplinary Reviews: Data Mining and Knowledge
  Discovery 10(6):e1381

\bibitem[{G{\"o}z{\"u}a{\c{c}}{\i}k and Can(2021)}]{gozuaccik2021concept}
G{\"o}z{\"u}a{\c{c}}{\i}k {\"O}, Can F (2021) Concept learning using one-class
  classifiers for implicit drift detection in evolving data streams. Artificial
  Intelligence Review 54(5):3725--3747

\bibitem[{G{\"o}z{\"u}a{\c{c}}{\i}k et~al(2019)G{\"o}z{\"u}a{\c{c}}{\i}k,
  B{\"u}y{\"u}k{\c{c}}ak{\i}r, Bonab, and Can}]{gozuaccik2019unsupervised}
G{\"o}z{\"u}a{\c{c}}{\i}k {\"O}, B{\"u}y{\"u}k{\c{c}}ak{\i}r A, Bonab H, et~al
  (2019) Unsupervised concept drift detection with a discriminative classifier.
  In: Proceedings of the 28th ACM International Conference on Information and
  Knowledge Management, pp 2365--2368

\bibitem[{Guo and Gu(2011)}]{guo2011multi}
Guo Y, Gu S (2011) Multi-label classification using conditional dependency
  networks. In: Twenty-Second International Joint Conference on Artificial
  Intelligence

\bibitem[{Hammami et~al(2020)Hammami, Sayed-Mouchaweh, Mouelhi, and
  Ben~Said}]{hammami2020neural}
Hammami Z, Sayed-Mouchaweh M, Mouelhi W, et~al (2020) Neural networks for
  online learning of non-stationary data streams: a review and application for
  smart grids flexibility improvement. Artificial Intelligence Review
  53:6111--6154

\bibitem[{Hoeffding(1963)}]{hoeffding1963}
Hoeffding W (1963) Probability inequalities for sums of bounded random
  variables. Journal of the American Statistical Association 58(301):13--30.
  \doi{10.1080/01621459.1963.10500830}

\bibitem[{Iwashita and Papa(2018)}]{iwashita2018overview}
Iwashita AS, Papa JP (2018) An overview on concept drift learning. Ieee Access
  7:1532--1547

\bibitem[{John(1995)}]{john1995estimating}
John G (1995) Estimating continuous distributions in bayesian classifiers. In:
  Proc. 11th Conference on Uncertainty in Artificial Intelligence

\bibitem[{Kendall(1938)}]{kendall1938new}
Kendall MG (1938) A new measure of rank correlation. Biometrika 30(1/2):81--93

\bibitem[{Koh(2016)}]{koh2016cd}
Koh YS (2016) Cd-tds: Change detection in transactional data streams for
  frequent pattern mining. In: 2016 International Joint Conference on Neural
  Networks (IJCNN), IEEE, pp 1554--1561

\bibitem[{Lu et~al(2018)Lu, Liu, Dong, Gu, Gama, and Zhang}]{lu2018learning}
Lu J, Liu A, Dong F, et~al (2018) Learning under concept drift: A review. IEEE
  Transactions on Knowledge and Data Engineering 31(12):2346--2363

\bibitem[{de~Mello et~al(2019)de~Mello, Vaz, Grossi, and
  Bifet}]{de2019learning}
de~Mello RF, Vaz Y, Grossi CH, et~al (2019) On learning guarantees to
  unsupervised concept drift detection on data streams. Expert Systems with
  Applications 117:90--102

\bibitem[{Montiel et~al(2018)Montiel, Read, Bifet, and
  Abdessalem}]{montiel2018scikit}
Montiel J, Read J, Bifet A, et~al (2018) Scikit-multiflow: A multi-output
  streaming framework. The Journal of Machine Learning Research
  19(1):2915--2914

\bibitem[{Nam et~al(2017)Nam, Menc{\'\i}a, Kim, and
  F{\"u}rnkranz}]{nam2017maximizing}
Nam J, Menc{\'\i}a EL, Kim HJ, et~al (2017) Maximizing subset accuracy with
  recurrent neural networks in multi-label classification. In: Proceedings of
  the 31st International Conference on Neural Information Processing Systems,
  pp 5419--5429

\bibitem[{Nuray and Can(2006)}]{nuray2006automatic}
Nuray R, Can F (2006) Automatic ranking of information retrieval systems using
  data fusion. Information Processing \& Management 42(3):595--614

\bibitem[{Pears et~al(2014)Pears, Sakthithasan, and Koh}]{pears2014detecting}
Pears R, Sakthithasan S, Koh YS (2014) Detecting concept change in dynamic data
  streams. Machine Learning 97(3):259--293

\bibitem[{Pedronette and Torres(2015)}]{pedronette2015unsupervised}
Pedronette DCG, Torres RdS (2015) Unsupervised effectiveness estimation for
  image retrieval using reciprocal rank information. In: 2015 28th SIBGRAPI
  Conference on Graphics, Patterns and Images, IEEE, pp 321--328

\bibitem[{Pesaranghader and Viktor(2016)}]{pesaranghader2016fast}
Pesaranghader A, Viktor HL (2016) Fast hoeffding drift detection method for
  evolving data streams. In: Joint European Conference on Machine Learning and
  Knowledge Discovery in Databases, Springer, pp 96--111

\bibitem[{Pesaranghader et~al(2018{\natexlab{a}})Pesaranghader, Viktor, and
  Paquet}]{pesaranghader2018reservoir}
Pesaranghader A, Viktor H, Paquet E (2018{\natexlab{a}}) Reservoir of diverse
  adaptive learners and stacking fast hoeffding drift detection methods for
  evolving data streams. Machine Learning 107(11):1711--1743

\bibitem[{Pesaranghader et~al(2018{\natexlab{b}})Pesaranghader, Viktor, and
  Paquet}]{pesaranghader2018mcdiarmid}
Pesaranghader A, Viktor HL, Paquet E (2018{\natexlab{b}}) Mcdiarmid drift
  detection methods for evolving data streams. In: 2018 International Joint
  Conference on Neural Networks (IJCNN), IEEE, pp 1--9

\bibitem[{Pinag{\'e} et~al(2020)Pinag{\'e}, dos Santos, and
  Gama}]{pinage2020drift}
Pinag{\'e} F, dos Santos EM, Gama J (2020) A drift detection method based on
  dynamic classifier selection. Data Mining and Knowledge Discovery
  34(1):50--74

\bibitem[{Pintas et~al(2021)Pintas, Fernandes, and Garcia}]{pintas2021feature}
Pintas JT, Fernandes LA, Garcia ACB (2021) Feature selection methods for text
  classification: a systematic literature review. Artificial Intelligence
  Review pp 1--52

\bibitem[{Pukelsheim(1994)}]{pukelsheim1994three}
Pukelsheim F (1994) The three sigma rule. The American Statistician
  48(2):88--91

\bibitem[{Raab et~al(2020)Raab, Heusinger, and Schleif}]{raab2020reactive}
Raab C, Heusinger M, Schleif FM (2020) Reactive soft prototype computing for
  concept drift streams. Neurocomputing

\bibitem[{Read et~al(2011)Read, Pfahringer, Holmes, and
  Frank}]{read2011classifier}
Read J, Pfahringer B, Holmes G, et~al (2011) Classifier chains for multi-label
  classification. Machine Learning 85(3):333

\bibitem[{Read et~al(2016)Read, Reutemann, Pfahringer, and
  Holmes}]{read2016meka}
Read J, Reutemann P, Pfahringer B, et~al (2016) Meka: a
  multi-label/multi-target extension to weka. The Journal of Machine Learning
  Research 17(1):667--671

\bibitem[{dos Reis et~al(2016)dos Reis, Flach, Matwin, and
  Batista}]{dos2016fast}
dos Reis DM, Flach P, Matwin S, et~al (2016) Fast unsupervised online drift
  detection using incremental kolmogorov-smirnov test. In: Proceedings of the
  22nd ACM SIGKDD International Conference on Knowledge Discovery and Data
  Mining, pp 1545--1554

\bibitem[{Roseberry and Cano(2018)}]{roseberry2018multi}
Roseberry M, Cano A (2018) Multi-label knn classifier with self adjusting
  memory for drifting data streams. In: Second International Workshop on
  Learning with Imbalanced Domains: Theory and Applications, PMLR, pp 23--37

\bibitem[{Sa{\l}abun and Urbaniak(2020)}]{salabun2020new}
Sa{\l}abun W, Urbaniak K (2020) A new coefficient of rankings similarity in
  decision-making problems. In: International Conference on Computational
  Science, Springer, pp 632--645

\bibitem[{Sethi and Kantardzic(2017)}]{sethi2017reliable}
Sethi TS, Kantardzic M (2017) On the reliable detection of concept drift from
  streaming unlabeled data. Expert Systems with Applications 82:77--99

\bibitem[{Shi et~al(2014)Shi, Wen, Feng, and Zhao}]{shi2014drift}
Shi Z, Wen Y, Feng C, et~al (2014) Drift detection for multi-label data streams
  based on label grouping and entropy. In: 2014 IEEE International Conference
  on Data Mining Workshop, IEEE, pp 724--731

\bibitem[{Spearman(1987)}]{spearman1987proof}
Spearman C (1987) The proof and measurement of association between two things.
  The American Journal of Psychology 100(3/4):441--471

\bibitem[{Subhashini et~al(2021)Subhashini, Li, Zhang, Atukorale, and
  Wu}]{subhashini2021mining}
Subhashini L, Li Y, Zhang J, et~al (2021) Mining and classifying customer
  reviews: a survey. Artificial Intelligence Review pp 1--47

\bibitem[{Tsoumakas and Katakis(2007)}]{tsoumakas2007multi}
Tsoumakas G, Katakis I (2007) Multi-label classification: An overview.
  International Journal of Data Warehousing and Mining (IJDWM) 3(3):1--13

\bibitem[{Veloso et~al(2021)Veloso, Gama, Malheiro, and
  Vinagre}]{veloso2021hyperparameter}
Veloso B, Gama J, Malheiro B, et~al (2021) Hyperparameter self-tuning for data
  streams. Information Fusion 76:75--86

\bibitem[{Vigna(2015)}]{vigna2015weighted}
Vigna S (2015) A weighted correlation index for rankings with ties. In:
  Proceedings of the 24th International Conference on World Wide Web, pp
  1166--1176

\bibitem[{Wang and Zhang(2020)}]{Wang_2020_CVPR}
Wang D, Zhang S (2020) Unsupervised person re-identification via multi-label
  classification. In: Proceedings of the IEEE/CVF Conference on Computer Vision
  and Pattern Recognition (CVPR)

\bibitem[{Wang et~al(2016)Wang, Yang, Mao, Huang, Huang, and Xu}]{wang2016cnn}
Wang J, Yang Y, Mao J, et~al (2016) Cnn-rnn: A unified framework for
  multi-label image classification. In: Proceedings of the IEEE Conference on
  Computer Vision and Pattern Recognition, pp 2285--2294

\bibitem[{Wang et~al(2020)Wang, Jin, and Fehringer}]{wang2020concept}
Wang P, Jin N, Fehringer G (2020) Concept drift detection with false positive
  rate for multi-label classification in iot data stream. In: 2020
  International Conference on UK-China Emerging Technologies (UCET), IEEE, pp
  1--4

\bibitem[{Xu et~al(2019)Xu, Shi, Tsang, Ong, Gong, and Shen}]{xu2019survey}
Xu D, Shi Y, Tsang IW, et~al (2019) Survey on multi-output learning. IEEE
  Transactions on Neural Networks and Learning Systems

\bibitem[{Xue et~al(2011)Xue, Zhang, Zhang, Wu, Fan, and
  Lu}]{xue2011correlative}
Xue X, Zhang W, Zhang J, et~al (2011) Correlative multi-label multi-instance
  image annotation. In: 2011 International Conference on Computer Vision, IEEE,
  pp 651--658

\bibitem[{Zhang and Zhang(2010)}]{zhang2010multi}
Zhang ML, Zhang K (2010) Multi-label learning by exploiting label dependency.
  In: Proceedings of the 16th ACM SIGKDD International Conference on Knowledge
  Discovery and Data Mining, pp 999--1008

\bibitem[{Zhang and Zhou(2013)}]{zhang2013review}
Zhang ML, Zhou ZH (2013) A review on multi-label learning algorithms. IEEE
  Transactions on Knowledge and Data Engineering 26(8):1819--1837

\bibitem[{Zheng et~al(2019)Zheng, Li, Chu, and Hu}]{zheng2019survey}
Zheng X, Li P, Chu Z, et~al (2019) A survey on multi-label data stream
  classification. IEEE Access

\bibitem[{{\v{Z}}liobaite(2010)}]{vzliobaite2010change}
{\v{Z}}liobaite I (2010) Change with delayed labeling: When is it detectable?
  In: 2010 IEEE International Conference on Data Mining Workshops, IEEE, pp
  843--850

\end{thebibliography}

\begin{appendices}
\onecolumn
\section{Detailed Calculation Example}\label{secA1}

Assume that a given data stream $S_{0,t}={d_0,...,d_t}$, in a time window $[0, t]$, where $d_i=(X_i,y_i)$, the window size $w = 3$ and three classes, the most recent six predictions are:
\begin{equation}
    S = ((0,0,1),(1,1,1),(0,1,1), \mathbf{ (1,0,1),(1,1,0),(1,0,1))} 
\end{equation}

We first calculate two co-occurrence matrices $M_{new}$ and $M_{old}$. For $M_{new}$ we use the most recent $w$ labels which are highlighted in bold. The next $w$ labels after that are used for $M_{old}$. The matrices are calculated by counting the number of times each label occurs together. For instance, for the third sample $(0,1,1)$, the second and third labels are true together, so $M_{old}(2,3)$ and $M_{old}(3,2)$ are incremented by one. Each row of the matrices represents the co-occurrence for a different label. For the given stream, the resulting matrices are the following: 
\renewcommand\arraystretch{1.5}

\begin{equation*}
\begin{aligned}[c]
M_{old} = \begin{bmatrix}
                0&\quad1&\quad1 \\
                1&\quad0&\quad2 \\
                1&\quad2&\quad0
              \end{bmatrix} 
\end{aligned} \tag{A2}
\qquad \qquad \qquad \qquad \qquad \qquad \qquad \qquad 
\begin{aligned}[c]
M_{new} = \begin{bmatrix}
                0&\quad1&\quad2 \\
                1&\quad0&\quad0 \\
                2&\quad0&\quad0
              \end{bmatrix} 
\end{aligned} 
\end{equation*}
\\

Next, we calculate the ranking of the co-occurrences for each label, i.e, which labels occur more frequently together with the currently ranked label. A.4 is for $M_{old}$ and A.5 is for $M_{new}$, where $L_i$ represents the rankings for each past predicted label and $L^{\prime}_{i}$ is used for the most recently predicted labels.


\begin{equation*}
\begin{aligned}[c]
\begin{split}
    L_1 \leftarrow l_2=l_3 \\
    L_2 \leftarrow l_3>l_1\\
    L_3 \leftarrow l_2>l_1
  \end{split} 
\end{aligned} \tag{A3}
\qquad \qquad \qquad \qquad \qquad \qquad \qquad \qquad 
\begin{aligned}[c]
\begin{split}
    L^{\prime}_{1} \leftarrow l_{3}>l_{2} \\
    L^{\prime}_{2} \leftarrow l_{1} > l_{3} \\
    L^{\prime}_{3} \leftarrow l_{1}>l_{2}
    \end{split}
\end{aligned} 
\end{equation*}
\\

We perform reciprocal rank fusion on the obtained rankings to get a global, aggregated ranking for the old and new labels. For tied rankings, we assume they each have the same ranking. $r_i$ and $r^{\prime}_i$ are the reciprocal rank of each label for old and new predictions in their respective order. The calculated ranks are sorted in ascending order to get the global rankings $R_{old}$ and $R_{new}$. If two reciprocal ranks are equal, they are sorted based on their label indexes.


\begin{equation*}
\begin{aligned}[c]
\begin{split}
    r_1 = \frac{1}{\frac{1}{2} + \frac{1}{2}} = 1 \\
    r_2 = \frac{1}{\frac{1}{1} + \frac{1}{1}} = \frac{1}{2}\\
    r_3 = \frac{1}{\frac{1}{1} + \frac{1}{1}} = \frac{1}{2} \\ \\
    R_{old} = [l_2, l_3, l_1]
  \end{split} 
\end{aligned} \tag{A4}
\qquad \qquad \qquad \qquad \qquad \qquad \qquad \qquad 
\begin{aligned}[c]
\begin{split}
    r^{\prime}_{1} = \frac{1}{\frac{1}{1} + \frac{1}{1}} = \frac{1}{2} \\
    r^{\prime}_{2} = \frac{1}{\frac{1}{2} + \frac{1}{2}} = 1 \\
    r^{\prime}_{3} = \frac{1}{\frac{1}{1} + \frac{1}{2}} = \frac{2}{3} \\ \\
    R_{new} = [l_1, l_3, l_2]
    \end{split}
\end{aligned} 
\end{equation*}
\\

The obtained $R_{old}$ and $R_{new}$ rankings are then used to calculate the WS coefficient to measure the rank correlation between the two rankings. In A.8, the ranks represent the ranking position of the \emph{i-th} label where $R^{old} = [2, 0, 1]$ and $R^{new} = [0, 2, 1]$. 

\begin{equation}
        C = 1 - \sum_{i=1}^{3}\bigg(2^{-R^{new}_{i}}\cdot \frac{\vert R^{new}_{i} - R^{old}_{i}\vert}{max\{\vert 1-R^{new}_{i}\vert,\vert 3-R^{new}_{i}\vert\}}\bigg) \tag{A5}
\end{equation}

\begin{equation}
\begin{split}
        C = 1 - \bigg(2^{0} \cdot \frac{\vert 0-2\vert}{max\{\vert 1-0\vert,\vert 3-0\vert\}} + 2^{-2} \cdot \frac{\vert 2-0\vert}{max\{\vert 1-2\vert,\vert 3-2\vert\}} + \\ 2^{-1} \cdot \frac{\vert 1-1\vert}{max\{\vert 1-1\vert,\vert 3-1\vert\}} \bigg)
\end{split} \tag{A6}
\end{equation}

\begin{equation}
        C = 1 - \bigg(\frac{2}{3} + \frac{1}{2} + 0 \bigg) = -0.167 \tag{A7}
\end{equation}

For the remaining part, if the obtained correlation $C$ is less than $\mu - t\sigma$, where $\mu$ is the mean, $\sigma$ is the standard deviation and $t$ is the $\sigma$ multiplier used in three sigma rule, the obtained correlation is added to an anomaly list. The drift is detected when the length of the anomaly list generated from $W_{corr}$ is greater than the set $L$ threshold.

\end{appendices}



\end{document}